\documentclass[10pt,twocolumn,letterpaper]{article}

\usepackage[pagenumbers]{iccv} 

\definecolor{iccvblue}{rgb}{0.21,0.49,0.74}
\usepackage[pagebackref,breaklinks,colorlinks,allcolors=iccvblue]{hyperref}

\usepackage{hyperref}       
\usepackage{url}            
\usepackage{booktabs}       
\usepackage{nicefrac}       
\usepackage{microtype}      

\usepackage{latexsym,setspace,mathtools,bm}
\usepackage{subcaption}
\usepackage{float}
\usepackage{array}
\usepackage{comment}

\usepackage{algorithm}
\usepackage{algpseudocode}
\usepackage{multicol}
\usepackage{lipsum}
\usepackage{xcolor}
\usepackage{colortbl}
\usepackage{graphicx}
\usepackage{array}
\usepackage{svg}
\usepackage{float}
\usepackage{booktabs}
\usepackage{multirow}
\usepackage{listings}
\usepackage{amsmath}
\usepackage{hyperref}
\usepackage{float}

\lstdefinelanguage{Python}{
    keywords={def, return, if, else, for, in, while, import, from, as, None, True, False, class, with, try, except},
    keywordstyle=\color{blue}\bfseries,
    ndkeywordstyle=\color{teal},
    identifierstyle=\color{black},
    sensitive=true,
    comment=[l]{\#},
    commentstyle=\color{gray},
    stringstyle=\color{green!50!black},
    morestring=[b]',
    morestring=[b]"
}

\lstdefinestyle{custom_python}{
    language=Python,
    basicstyle=\ttfamily\footnotesize,
    keywordstyle=\color{nice_red}\bfseries,
    stringstyle=\color{green!50!black},
    commentstyle=\color{dark_green},
    identifierstyle=\color{black},
    numbers=left,
    numberstyle=\tiny\color{gray},
    stepnumber=1,
    numbersep=5pt,
    showspaces=false,
    showstringspaces=false,
    showtabs=false,
    frame=single,
    rulecolor=\color{black},
    tabsize=4,
    breaklines=true,
    breakatwhitespace=true,
    captionpos=b,
    moredelim=[is][\bfseries\color{brown}]{@}{@} 
}

\definecolor{nice_blue}{RGB}{65, 105, 225}
\definecolor{nice_red}{RGB}{168, 34, 34}
\definecolor{IGCBlue}{HTML}{16197A}
\definecolor{dark_green}{RGB}{20, 110, 10}

\definecolor{graph_blue}{RGB}{144, 195, 212}
\definecolor{graph_purple}{RGB}{195, 144, 212}
\definecolor{graph_green}{RGB}{161, 212, 144}
\definecolor{graph_orred}{RGB}{212, 161, 144}
\definecolor{lavender}{rgb}{0.9, 0.9, 0.98}
\definecolor{weakgray}{gray}{0.6}

\usepackage{graphicx}

\def\paperID{14867} 
\def\confName{ICCV}
\def\confYear{2025}

\title{Training Noise Token Pruning}

\author{Mingxing Rao, Bohan Jiang, Daniel Moyer\\
Vanderbilt University\\
Nashville, TN 37235,USA\\
{\tt\small \{mingxing.rao, bohan.jiang, daniel.moyer\}@vanderbilt.edu}
}

\begin{document}
\maketitle
\begin{abstract}
In the present work we present Training Noise Token (TNT) Pruning for vision transformers. Our method relaxes the discrete token dropping condition to continuous additive noise, providing smooth optimization in training, while retaining discrete dropping computational gains in deployment settings. We provide theoretical connections to Rate-Distortion literature, and empirical evaluations on the ImageNet dataset using ViT and DeiT architectures demonstrating TNT's advantages over previous pruning methods. 
\end{abstract}    
\begin{figure*}[t]
    \centering
    \includegraphics[width=\textwidth]{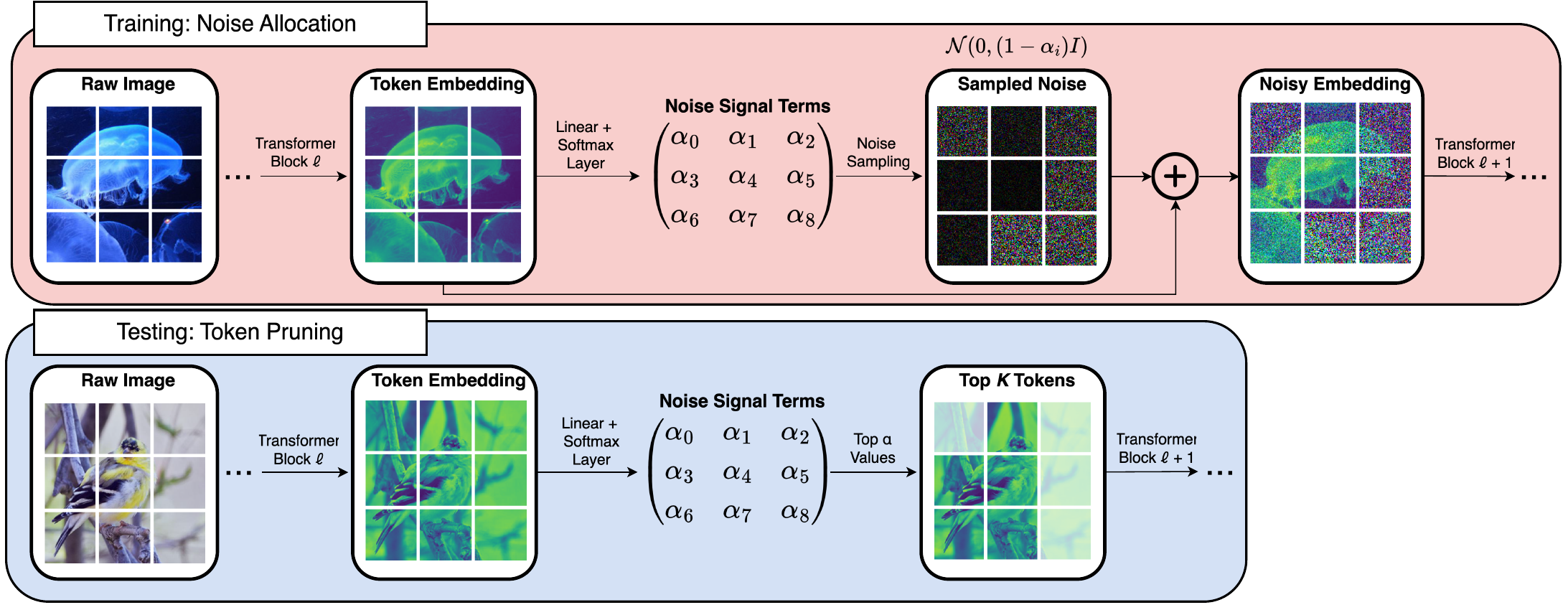}
    \caption{\textbf{Training Noise Token Pruning (TNT).} Our proposed method computes a relevance term $\alpha_i$ for each token. In training (diagrammed at top), these terms dictate an amount of noise added to the token, while at test time they indicate pruning order.   
    }
    \label{fig:mainfig}
\end{figure*}

\section{Introduction}
\label{sec:intro}


Token pruning is a class of methods for reducing computational load in transformers \cite{vaswani2017attention} by reducing the input length. While transformers are already somewhat robust to dropping a small number of tokens at random, learned dropping schemes enable larger dropping rates and thereby higher speed-ups with smaller accuracy penalties. These gains are only amplified by operations super-linear cost scaling (e.g., attention) and memory footprint.

Token pruning methods exploit predictive information differences between tokens; for a given task, some tokens are more useful than others. By dropping the least informative tokens first, a learned method can preserve the most amount of accuracy while removing desired number of tokens. While these methods were originally explored in a natural language processing context, token redundancy is arguably stronger in image transformers, and multiple token pruning methods for vision models (e.g. ViT \cite{dosovitskiy2020image}) have been proposed \cite{rao2021dynamicvit,yin2022vit,wang2024zero,xu2022evo,liang2022not}.

Token relevance in many transformer models is already captured by the \texttt{CLS} token, and the attention values it attributes to the other tokens. The \texttt{CLS} tokens are directly predictive of their given task, often trained using the appropriate prediction loss, mapped to the label domain by a single linear layer. Thus, the tokens that the \texttt{CLS} token attends to are the relevant tokens for that task.
This intuition is implemented in \cite{liang2022not, fayyaz2022adaptive}; moreover, Haurum et al. \cite{haurum2023tokens} show that simply ranking tokens by \texttt{CLS} attention score and taking the desired number (the ``Top-K'' method) outperforms most methods proposed to date. However, these methods are conditional on the \texttt{CLS} token existing in the architecture, and the primary task matching the \texttt{CLS} token training task.





In this paper, we provide high fidelity estimates of relevance without the \texttt{CLS} token by using a variational Information Bottleneck \cite{tishby2000information,alemi2016deep}. The VIB optimization objective is a trade-off between transmission rate (the number of bits in a message) and relevance to some extrinsic factor. This is exactly the same trade-off in Token Dropping, where we would like minimal token counts (analogous to transmission rate) for maximal relevance. We demonstrate that our proposed method has superior performance in comparison to recent pruning methods using either stochastic discrete dropping and matches or exceeds the performance of \texttt{CLS}-attention methods even without the \texttt{CLS} token, both in terms of accuracy and in computational cost.

In the present work we provide:
\begin{itemize}
\item A novel method for token-pruning that provides state-of-the-art performance with respect to the accuracy/computation load trade-off.
\item A justification and intuition for that method based on the information bottleneck.
\item Empirical experiments demonstrating the use and utility of our method along with previous methods as baselines, with evaluations on ImageNet \cite{dosovitskiy2020image} using two common image transformers, ViT \cite{dosovitskiy2020image} and DeiT \cite{touvron2021training}, as base architectures.
\end{itemize}
Our code can be found at \url{https://github.com/mx-ethan-rao/tnt}.

\section{Related Work}
\label{sec:formatting}

\paragraph{Token Pruning:}
Token pruning (or ``token dropout'') has been explored for both transformer-based language models \cite{goyal2020power,kim2020length,kim2022learned} and vision transformers \cite{rao2021dynamicvit,xu2022evo,yin2022vit,wang2024zero}. Broadly speaking, these methods can be separated into two categories: stochastic dropout methods, where tokens are randomly removed based on a per-token computed likelihood \cite{goyal2020power,kim2020length,kim2022learned,rao2021dynamicvit,yin2022vit}, and heuristic attention-based methods \cite{liang2022not,wang2024zero}

Rao et al. 2021 \cite{rao2021dynamicvit} introduces Dynamic ViT, which is notable as the first token pruning method for vision transformer \cite{dosovitskiy2020image}. It is prototypical of the stochastic dropout family of methods: it defines a relaxation of the rate criterion and samples tokens according to an inclusion likelihood. Yin et al. 2022 \cite{yin2022vit} introduce a refinement of this method (Adaptive ViT), incorporating a halting module and associated score, as well as a loss function that encourages early stopping.

Along a different path, multiple heuristics have been introduced for token pruning based on the attention scores \cite{liang2022not,wang2024zero}. These use the intuition that tokens to which other tokens ascribe high attention (highly attended tokens) are of high importance. Liang et al. 2022 \cite{liang2022not} additionally merges the lowest attended tokens, while Wang et al. 2024 puts this both into a graph ranking framework (PageRank \cite{page1999pagerank}) and into the zero-shot context.

A simple baseline version of an attention-based heuristic, ``Top-K pruning'', was also found to perform competitively as well \cite{haurum2023tokens}. This method simply ranks tokens based on the attention distribution to a CLS token, and then truncates after the top $K$. While it is not directly applicable to architectures without CLS tokens or non-classification tasks, our results in Section \ref{sec:experiments} show that it outperforms most other methods where it can be applied.

\paragraph{Merger Methods:} Complementary to dropping methods, token mergers and similarity-based pruning methods also decrease transformer computational cost by reducing the size of the token set \cite{bolya2022token, wang2024zero, liang2022not}. Similarity-based pruning \cite{wang2024zero} exploits exactly the opposite problem in the token set; instead of removing low relevance tokens they generally merge redundant tokens. Merger methods are more general, also possibly allowing for encoding of background context variables \cite{bolya2022token}. The use of these methods is not mutually exclusive with token pruning \cite{wang2024zero}. While in the present work we do not provide a completely novel merger method, we provide improvements on an existing similarity-based pruning step, and provide justification for its necessity.

\paragraph{Information Bottleneck: } Our framework for token dropping relies upon theory originally explored in the Information Bottleneck \cite{tishby2000information,alemi2016deep}, which in turn is based upon Rate-Distortion theory \cite{blahut1972computation}. The information bottleneck characterizes encodings in terms of their relevance (the additive inverse of a distortion metric) and a rate constraint. We place token pruning's two relevant metrics into this context, which then provides a natural relaxation for the rate constraint.

\begin{figure}[h]
    \centering
    \includegraphics[width=0.8\columnwidth]{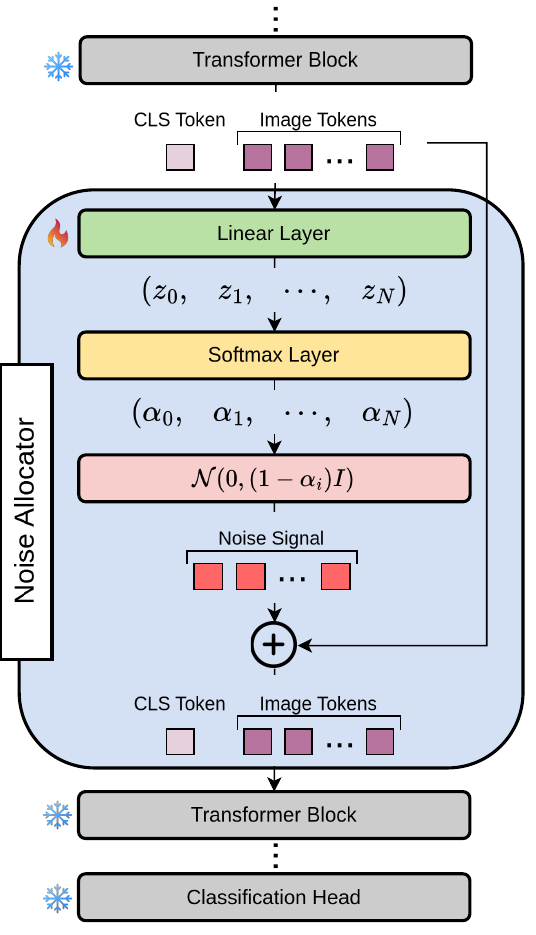}
    \caption{\textbf{Noise Allocator block architecture:} the block diagrammed above is injected into pre-trained models as a pruning layer. It takes the output of the previous Transformer block as input, then computes the noise signal terms \( \alpha \) using a linear layer followed by a \texttt{Softmax} function. During training it samples Gaussian noise conditioned on \( \alpha \) for each token, then adds the noise to the token embeddings. At test time, tokens are instead dropped. This pruning method can be trained with all parameters outside the noise allocator are frozen.}
    \label{fig:model_detail}
\end{figure}

\section{Methodology}

The Information Bottleneck \cite{tishby2000information} is a Rate-Distortion trade-off problem for transmitting a signal $s$ relevant to labels $y$ through an encoding $x$. In its setup, we're asked to find the minimal transmission rate, i.e., the minimal effective number of bits sent to $x$, which is the mutual information $I(s,x)$, subject to a maximal distortion of $x$'s relevance to $y$, which is measured by $I(x,y)$. This is written as a constrained optimization
\begin{align}
\min_{p(z|x)} I(s,x) \text{ subject to } I(x,y) > T.
\end{align}
This is often rewritten as an unconstrained optimization
\begin{align}
\min_{p(z|x)} I(s,x) - \lambda I(x,y)
\end{align}
where $\lambda$ is a function of the original constraint $T$. Dividing by $\lambda$, there is an equivalent objective $\frac{1}{\lambda}I(s,x) - I(x,y)$, we can then rewrite the optimization as a maximization of relevance (or a minimization of negative relevance) constrained by effective transmission rate:
\begin{align}
\min_{p(z|x)} -I(x,y) \text{ subject to } I(s,x) < \tilde{T}. \label{eq:rewritten-IB}
\end{align}
While Eq. \ref{eq:rewritten-IB} is not the original formulation of the bottleneck objective and is arrived at using elementary algebraic manipulations, it more clearly illustrates the analogy to the Token Pruning problem. We wish to limit the number of tokens transmitted (and subsequently processed, etc.) to some fixed number L, while maintaining the highest prediction accuracy, which is equivalent to minimizing the cross entropy $H(y|x)$, which itself is the objective $-I(x,y)$ up to constant $-H(y)$. 

We can approximate solutions to the information bottleneck using the deep variational method provided by Alemi et al. \cite{alemi2016deep}. Consider an architecture with L repeated transformer blocks possibly after an initial token embedding layer (for example, ViT has 12 identical blocks), where the $\ell$th Transformer block $f^{(\ell)}$ with input $x \in \mathbb{R}^{N \times D}$ has $N$ tokens with embedding dimension $D$, and individual $x^{(\ell)}_{i}$ tokens. For every block where we would like to perform pruning, we introduce a linear token relevance predictor with a weight matrix $W^{(\ell)} \in \mathbb{R}^{D \times 1} $. 

We then compute $\alpha^{(\ell)}_i$ as 
\begin{align}
\alpha_i^{\ell} = \text{Softmax}(W^{(\ell)} f^\ell(x^{(\ell)})) \label{eq:alpha}
\end{align}
We then compute noise variables $\eta_i^\ell$ as
\begin{align}
\eta_i^{\ell} = (1 - \alpha_i^{(\ell)}) \varepsilon,
\end{align}

where $\varepsilon \sim \mathcal{N}(0, \beta I_{D})$ (``the reparameterization trick'' \cite{kingma2013auto}). Here, $\beta$ is a hyper-parameter that controls the total amplitude of noise added.  The $\alpha^{(\ell)}_i$'s are our estimate of the most relevant tokens; if the network could output arbitrary values in [0,1] for $\alpha$, clearly the highest accuracy solution sends all $\alpha^{(\ell)}_i$'s to 1, adding zero noise and thus preserving all the signal. Due to the \texttt{Softmax}, instead this noise allocation is constrained to always add $\beta$ amount of noise. This forces some tokens to be dropped; the solution that has the highest prediction accuracy is the one that adds the least noise (i.e., has the highest $\alpha^{(\ell)}_i$) to the most predictive tokens. Thus, $\alpha^{(\ell)}_i$ is an approximation to the relevance of $x^{(\ell)}_i$.

At test time, instead of adding noise, we use the $\alpha^{(\ell)}_i$ to rank tokens in order of relevance. By removing all but the top ranked solutions (similar to what the ``Top-K'' method does for the \texttt{CLS} token attention scores).


Classical results in Information Theory state that the mutual information (channel capacity) for each token to its noised variant has an upper bound proportional to $\log(1 + P_{signal}/P_{noise})$, where $P_{signal}/P_{noise}$ is the ratio of the power (amplitude) of the token signal and to that of the noise \cite{cover1999elements} (the signal to noise ratio).  Due to the layer-wise normalization (\texttt{LayerNorm}) of many transformer architectures, $P_{signal}$ is necessarily bounded. This solution can also be directly mapped onto the Deep Variational Information Bottleneck by viewing our initial tokens $x^{(\ell)}_{i}$ as the mean of their latent embeddings, and our $(1-\alpha^{(\ell)}_i)$ as the element-wise standard deviations.

\subsection{Similarity-based Pruning by Random Partition (Redundancy Removal)}

Our chosen bottleneck approximation only considers elementwise approximation.
Both $I(x,y)$ and $I(s,x)$ measure interaction information; conditional on adding one token $x_i$, the information about $y$ in another token $x_j$ might be higher than the marginal information $I(x_j,y)$. These synergies are not limited to dyads, and can have arbitrarily high order. Further, redundancies, where $I(x_i,y) + I(x_j,y) > I((x_i,x_j),y)$ mean that we might include additional, unnecessary information.
Even though individual tokens might be highly relevant to the classification problem (i.e, individually $I(x_i^\ell,y)$ might be high), relevant tokens sharing a high amount of information with a kept token represents wasted capacity. This weakness is shared by \texttt{CLS}-token based removal methods, and in order to avoid this waste, we implement the same redundancy removal method. We discuss its limitations in Section \ref{sec:limitations}.


The method is as follows: Tokens are randomly divided into two groups. We then identify the closest matching token in one group for each token in the other, recording the similarity scores of each paired token based on their token embeddings. Next, we prune the top-$r$ most similar pairs and prune the associated tokens in the second group.

Our similarity-based pruning approach closely resembles that of Zero-TP~\cite{wang2024zero}, with two key modifications.
First, rather than using the Key values as the partitioning metric, we directly use \( x^{(l)} \) so that it can be applied directly after the Noise Allocator step.
Second, instead of sequentially partitioning tokens based on their importance scores, we apply a random partitioning strategy. Our ablation study suggests that random partitioning yields improved performance for the proposed model.

\begin{figure*}[t]
    \centering
    \includegraphics[width=0.9\textwidth]{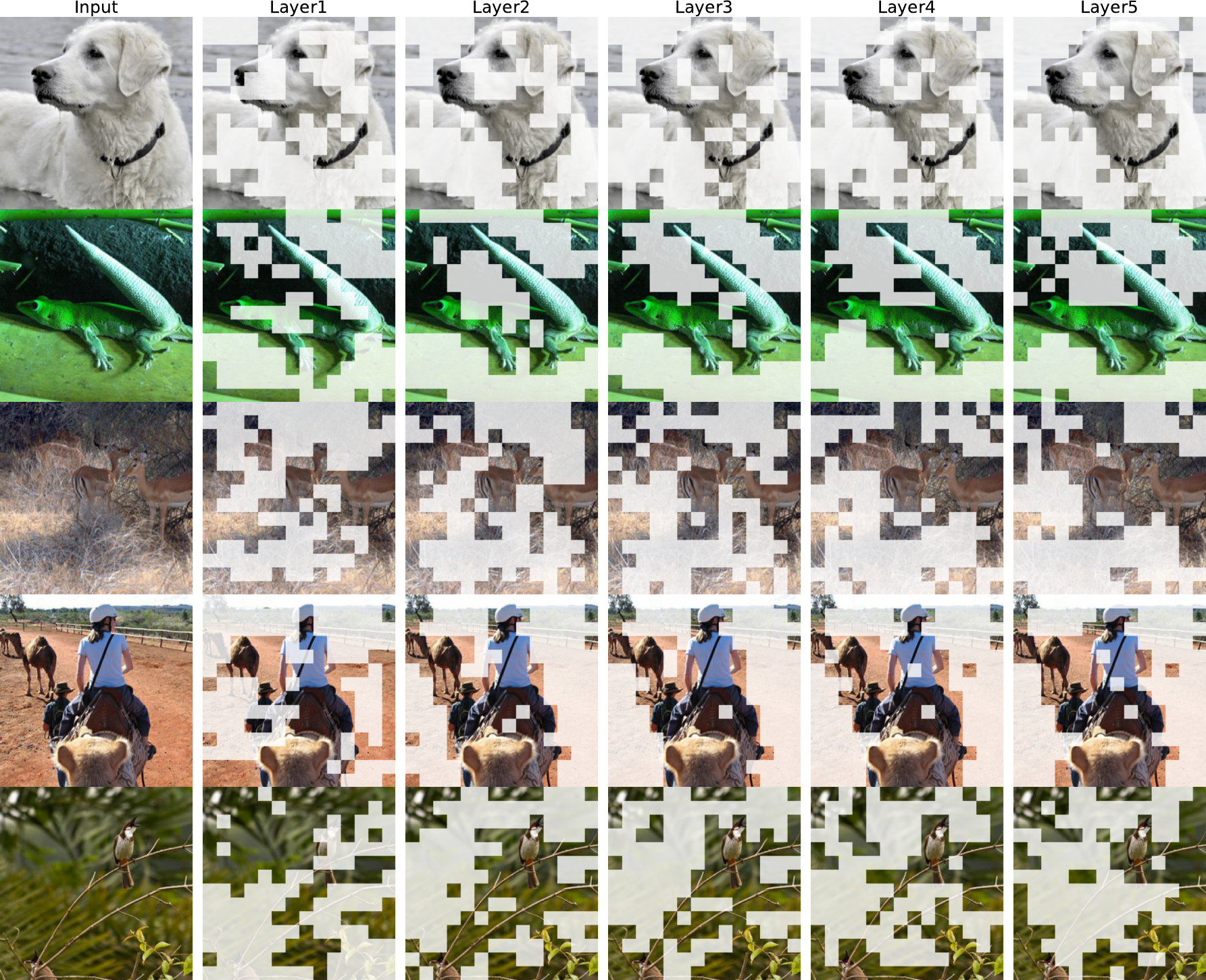}  
    \caption{Visualization of Token Pruning maps on ImageNet-1K: at \textbf{left} are the original images, and at each column \textbf{progressing right} are single layer prunings and their associated kept/dropped tokens, for layers 1-5 of the DeiT-B-Distil. model.}
    \label{fig:visualization}
\end{figure*}

\begin{figure*}[t]
    \centering
    \includegraphics[width=\textwidth]{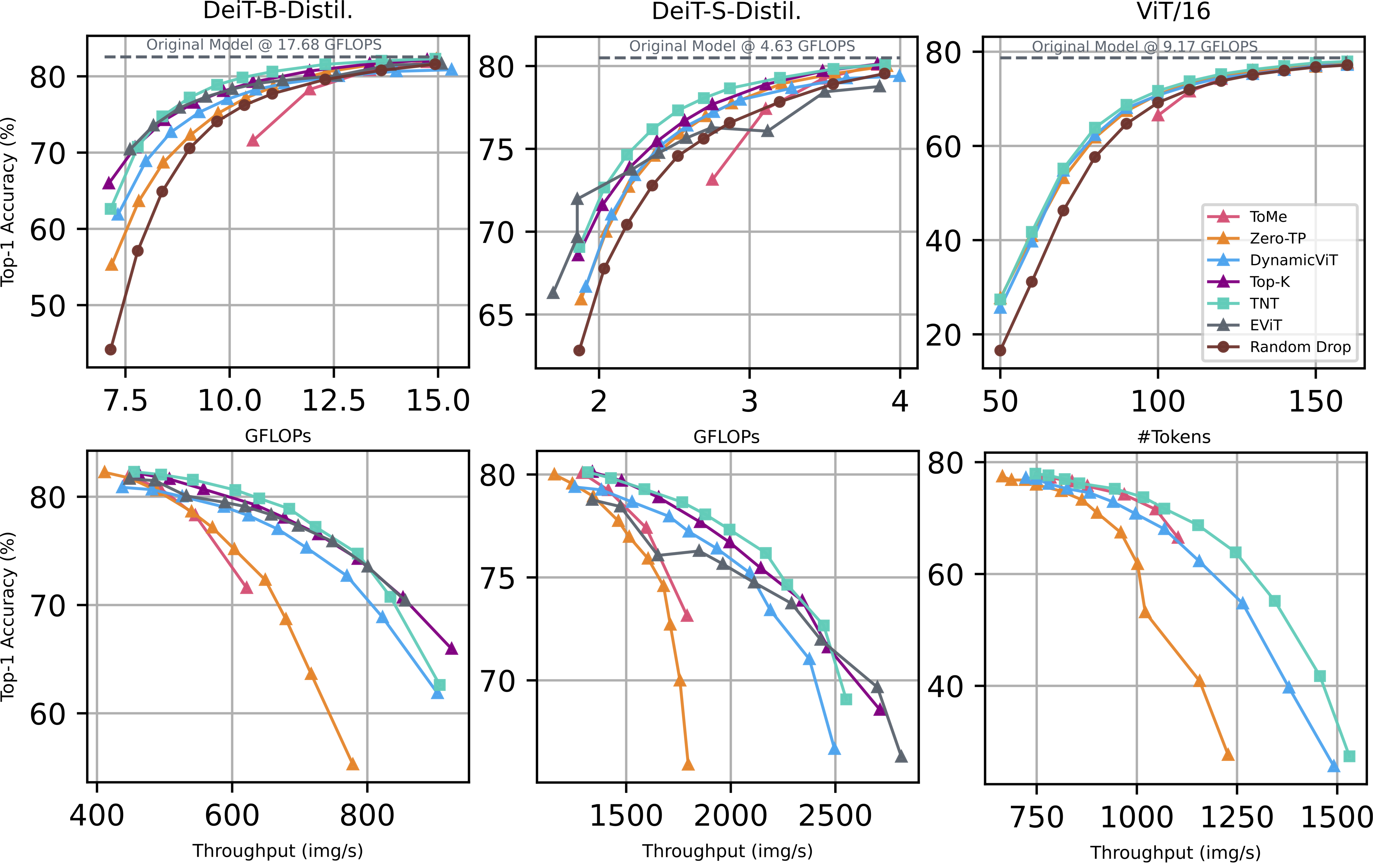}  
    \caption{\textbf{Single Layer Pruning results:} We plot the Top-1 Accuracy in the ImageNet-1k validation set for each of the pruning methods as a function of computational efficiency, in the \textbf{top row} measured by GFLOPs and in the \textbf{bottom row} measured by throughput,  for \textbf{single layer pruning}. The base model is DeiT-B-Distil in the \textbf{first column}, DeiT-S-Distil. in the \textbf{second column}, and ViT/16 in the \textbf{third column}.
    Note that the mean-pooled token embedding ViT in the \textbf{third column} has no \texttt{CLS} token, and thus EViT and Top-K cannot be applied to it.
    }
    \label{fig:single_layer}
\end{figure*}

\begin{table}[h!]
    \centering
    \setlength{\tabcolsep}{3pt} 
    \renewcommand{\arraystretch}{1.0} 
    \begin{tabular}{c|c|lcc|c}
        \hline
        \rowcolor{white}
        \multicolumn{2}{c}{} & \textbf{Method} & \textbf{Acc.} & \textbf{GFLOPs} & \textbf{TP (imgs/s)} \\
        \hline
        \multirow{16}{*}{\rotatebox[origin=c]{90}{High Token Regime}} & \multirow{6}{*}{\rotatebox[origin=c]{90}{$\tiny\text{GFLOPs } \text{\tiny $ \geq 8.0$}$}}
        & \textcolor{weakgray}{Deit-B-Distil.~\cite{touvron2021training}} & \textcolor{weakgray}{82.55} & \textcolor{weakgray}{17.68} & \textcolor{weakgray}{379} \\
        & & Top-K~\cite{haurum2023tokens} & 81.82 & 13.18 & 504 \\
        & & Zero-TP~\cite{wang2024zero} & 81.92 & 13.10 & 433 \\
        & & DynamicViT~\cite{rao2021dynamicvit} & 80.91 & 13.34 & 498 \\
        & & ToMe~\cite{bolya2022token} & 81.86 & 12.11 & 508 \\
        & & EViT~\cite{liang2022not} & 80.69 & 13.34 & 496 \\
        & & \cellcolor{lavender}TNT (ours) & \cellcolor{lavender}81.97 & \cellcolor{lavender}13.11 & \cellcolor{lavender}509 \\
        \cline{2-6}
        
        & \multirow{6}{*}{\rotatebox[origin=c]{90}{$\tiny\text{GFLOPs } \text{\tiny $ \geq 2.0$}$}}
        & \textcolor{weakgray}{Deit-S-Distil.~\cite{touvron2021training}} & \textcolor{weakgray}{80.49} & \textcolor{weakgray}{4.63} & \textcolor{weakgray}{1150} \\
        & & Top-K~\cite{haurum2023tokens} & 79.81 & 3.44 & 1496 \\
        & & EViT~\cite{liang2022not} & 79.16 & 3.48 & 1463 \\
        & & Zero-TP~\cite{wang2024zero} & 79.66 & 3.42 & 983 \\
        & & DynamicViT~\cite{rao2021dynamicvit} & 79.45 & 3.47 & 1448 \\
        & & \cellcolor{lavender}ToMe~\cite{bolya2022token} & \cellcolor{lavender}80.12 & \cellcolor{lavender}3.45 & \cellcolor{lavender}1281 \\
        & & TNT (ours) & 79.89 & 3.41 & 1443 \\
        \cline{2-6}
        
        & \multirow{4}{*}{\rotatebox[origin=c]{90}{$\tiny\text{GFLOPs } \text{\tiny $ \geq 4.0$}$}}
        & \textcolor{weakgray}{ViT/16~\cite{dosovitskiy2020image}} & \textcolor{weakgray}{78.70} & \textcolor{weakgray}{9.17} & \textcolor{weakgray}{644} \\
        & & Zero-TP~\cite{wang2024zero} & 77.07 & 6.75 & 688 \\
        & & ToMe~\cite{bolya2022token} & 77.03 & 6.97 & 734 \\
        & & DynamicViT~\cite{rao2021dynamicvit} & 78.56 & 7.23 & 830 \\
        & & \cellcolor{lavender}TNT (ours) & \cellcolor{lavender}77.32 & \cellcolor{lavender}6.73 & \cellcolor{lavender}842 \\
        
        \hline
        \multirow{16}{*}{\rotatebox[origin=c]{90}{Low Token Regime}} & \multirow{6}{*}{\rotatebox[origin=c]{90}{$\tiny\text{GFLOPs } \text{\tiny $ < 8.0$}$}}
        & \textcolor{weakgray}{Deit-B-Distil.~\cite{touvron2021training}} & \textcolor{weakgray}{82.55} & \textcolor{weakgray}{17.68} & \textcolor{weakgray}{379} \\
        & & Top-K~\cite{haurum2023tokens} & 55.98 & 5.93 & 1084 \\
        & & Zero-TP~\cite{wang2024zero} & 19.76 & 6.18 & 731 \\
        & & DynamicViT~\cite{rao2021dynamicvit} & 11.20 & 5.79 & 1102 \\
        & & ToMe~\cite{bolya2022token} & 43.37 & 5.89 & 995 \\
        & & \cellcolor{lavender}TNT (ours) & \cellcolor{lavender}59.93 & \cellcolor{lavender}5.87 & \cellcolor{lavender}1095 \\
        \cline{2-6}
        
        & \multirow{6}{*}{\rotatebox[origin=c]{90}{$\tiny\text{GFLOPs } \text{\tiny $ < 2.0$}$}}
        & \textcolor{weakgray}{Deit-S-Distil.~\cite{touvron2021training}} & \textcolor{weakgray}{80.49} & \textcolor{weakgray}{4.63} & \textcolor{weakgray}{1150} \\
        & & Top-K~\cite{haurum2023tokens} & 58.20 & 1.57 & 3003 \\
        & & Zero-TP~\cite{wang2024zero} & 27.75 & 1.63 & 1485 \\
        & & DynamicViT~\cite{rao2021dynamicvit} & 24.51 & 1.52 & 3037 \\
        & & ToMe~\cite{bolya2022token} & 62.88 & 1.55 & 2486 \\
        & & \cellcolor{lavender}TNT (ours) & \cellcolor{lavender}63.82 & \cellcolor{lavender}1.53 & \cellcolor{lavender}2973 \\
        \cline{2-6}
        
        & \multirow{4}{*}{\rotatebox[origin=c]{90}{$\tiny\text{GFLOPs } \text{\tiny $ < 4.0$}$}}
        & \textcolor{weakgray}{ViT/16~\cite{dosovitskiy2020image}} & \textcolor{weakgray}{78.70} & \textcolor{weakgray}{9.17} & \textcolor{weakgray}{644} \\
        & & Zero-TP~\cite{wang2024zero} & 0.63 & 3.26 & 1179 \\
        & & ToMe~\cite{bolya2022token} & 6.30 & 3.26 & 1485 \\
        & & \cellcolor{lavender}TNT (ours) & \cellcolor{lavender}13.42 & \cellcolor{lavender}2.98 & \cellcolor{lavender}1786 \\
        
        \hline
    \end{tabular}

\caption{
Top-1 Accuracy on the ImageNet-1K validation set (\textbf{Acc.}) and computation cost measured by GFLOPs and throughput (\textbf{TP}), for DeiT-B-Distil., DeiT-S-Distil., and ViT/16, in two different computing regimes (``High'' and ``Low'', defined for each base model). 
}
    \label{tab:multi-layer}
\end{table}

\begin{table*}[t]
\centering
\begin{tabular}{lccccccccccc}
\toprule
\multirow{2}{*}{} & \multicolumn{2}{c|}
{$\textbf{K=1.0}$} & \multicolumn{2}{c|}
{$\textbf{K=0.8}$} & \multicolumn{2}{c|}{$\textbf{K=0.6}$} & \multicolumn{2}{c|}{$\textbf{K=0.5}$} & \multicolumn{2}{c}{$\textbf{K=0.25}$} \\
\cmidrule(lr){2-3} \cmidrule(lr){4-5} \cmidrule(lr){6-7} \cmidrule(lr){8-9} \cmidrule(lr){10-11}
\multirow{2}{*}{} & \multicolumn{2}{c|}{\textbf{\#tokens=196}} & \multicolumn{2}{c|}{\textbf{\#tokens=156}} & \multicolumn{2}{c|}{\textbf{\#tokens=127}} & \multicolumn{2}{c|}{\textbf{\#tokens=98}} & \multicolumn{2}{c}{\textbf{\#tokens=49}} \\
\cmidrule(lr){2-3} \cmidrule(lr){4-5} \cmidrule(lr){6-7} \cmidrule(lr){8-9} \cmidrule(lr){10-11}

{$\textbf{DeiT-S-Distil.}$} & \textbf{Acc.} & \textbf{GFLOPs} & \textbf{Acc.} & \textbf{GFLOPs} & \textbf{Acc.} & \textbf{GFLOPs} & \textbf{Acc.} & \textbf{GFLOPs} & \textbf{Acc.} & \textbf{GFLOPs} \\
\midrule
Random Drop & \textcolor{weakgray}{80.50} & \textcolor{weakgray}{4.63} & 79.54 & 3.90 & 77.83 & 3.20 & 76.57 & 2.87 & 67.77 & 2.03 \\
ToMe~\cite{bolya2022token} &  -- & --  & 80.07 & 3.85 & 77.40 & 3.11 & 73.13 & 2.75 & - & - \\
Zero-TP~\cite{wang2024zero} &  -- & --  & 80.00 & 3.91 & 78.94 & 3.21 & 77.74 & 2.88 & 69.99 & 2.05 \\
DynamicViT~\cite{rao2021dynamicvit} &  -- & --  & 79.39 & 3.99 & 78.66 & 3.28 & 77.96 & 2.94 & 71.03 & 2.08 \\
Top-K~\cite{haurum2023tokens} &  -- & --  & \cellcolor{lavender}80.12 & 3.85 & 78.89 & 3.11 & 77.69 & 2.75 & 68.57 & 1.86 \\
EViT~\cite{liang2022not} &  -- & --  & 78.77 & 3.86 & 76.07 & 3.27 & 76.28 & 2.74 & 69.67 & 1.86 \\
TNT (ours) &  -- & --  & 80.11 & 3.90 & \cellcolor{lavender}79.29 & 3.20 & \cellcolor{lavender}78.65 & 2.87 & \cellcolor{lavender}72.66 & 2.03 \\
\midrule
\multirow{2}{*}{} & \multicolumn{2}{c|}{\textbf{\#tokens=196}} & \multicolumn{2}{c|}{\textbf{\#tokens=160}} & \multicolumn{2}{c|}{\textbf{\#tokens=130}} & \multicolumn{2}{c|}{\textbf{\#tokens=100}} & \multicolumn{2}{c}{\textbf{\#tokens=60}} \\
\cmidrule(lr){2-3} \cmidrule(lr){4-5} \cmidrule(lr){6-7} \cmidrule(lr){8-9} \cmidrule(lr){10-11}
\textbf{ViT/16} & \textbf{Acc.} & \textbf{GFLOPs} & \textbf{Acc.} & \textbf{GFLOPs} & \textbf{Acc.} & \textbf{GFLOPs} & \textbf{Acc.} & \textbf{GFLOPs} & \textbf{Acc.} & \textbf{GFLOPs}\\
\midrule
Random Drop & \textcolor{weakgray}{78.70} & \textcolor{weakgray}{9.17} & 77.17 & 7.69 & 75.12 & 6.50 & 69.21 & 5.33 & 31.16 & 3.81 \\
ToMe~\cite{bolya2022token} &  -- & --  & 77.71 & 7.66 & 75.70 & 6.42 & 66.46 & 5.22 & - & - \\
Zero-TP~\cite{wang2024zero} &  -- & --  & 77.43 & 7.32 & 75.96 & 6.52 & 70.92 & 5.35 & 40.84 & 3.83 \\
DynamicViT~\cite{rao2021dynamicvit} &  -- & --  & 77.18 & 7.98 & 75.25 & 6.68 & 70.77 & 5.43 & 39.68 & 4.06 \\
TNT (ours) &  -- & --  & \cellcolor{lavender}77.94 & 7.70 & \cellcolor{lavender}76.22 & 6.50 & \cellcolor{lavender}71.69 & 5.33 & \cellcolor{lavender}41.75 & 3.81 \\
\bottomrule
\end{tabular}
\caption{
Top-1 Accuracy on the ImageNet-1K validation set (\textbf{Acc.}) and computational cost (GFLOPs) across methods for differing keep rates $K$, using a single layer of token pruning. The \textbf{left most column} are the performance and computational cost of the base architectures.
}
\label{tab:single_layer}
\end{table*}

\begin{figure*}[h!]
    \centering
    \includegraphics[width=\textwidth]{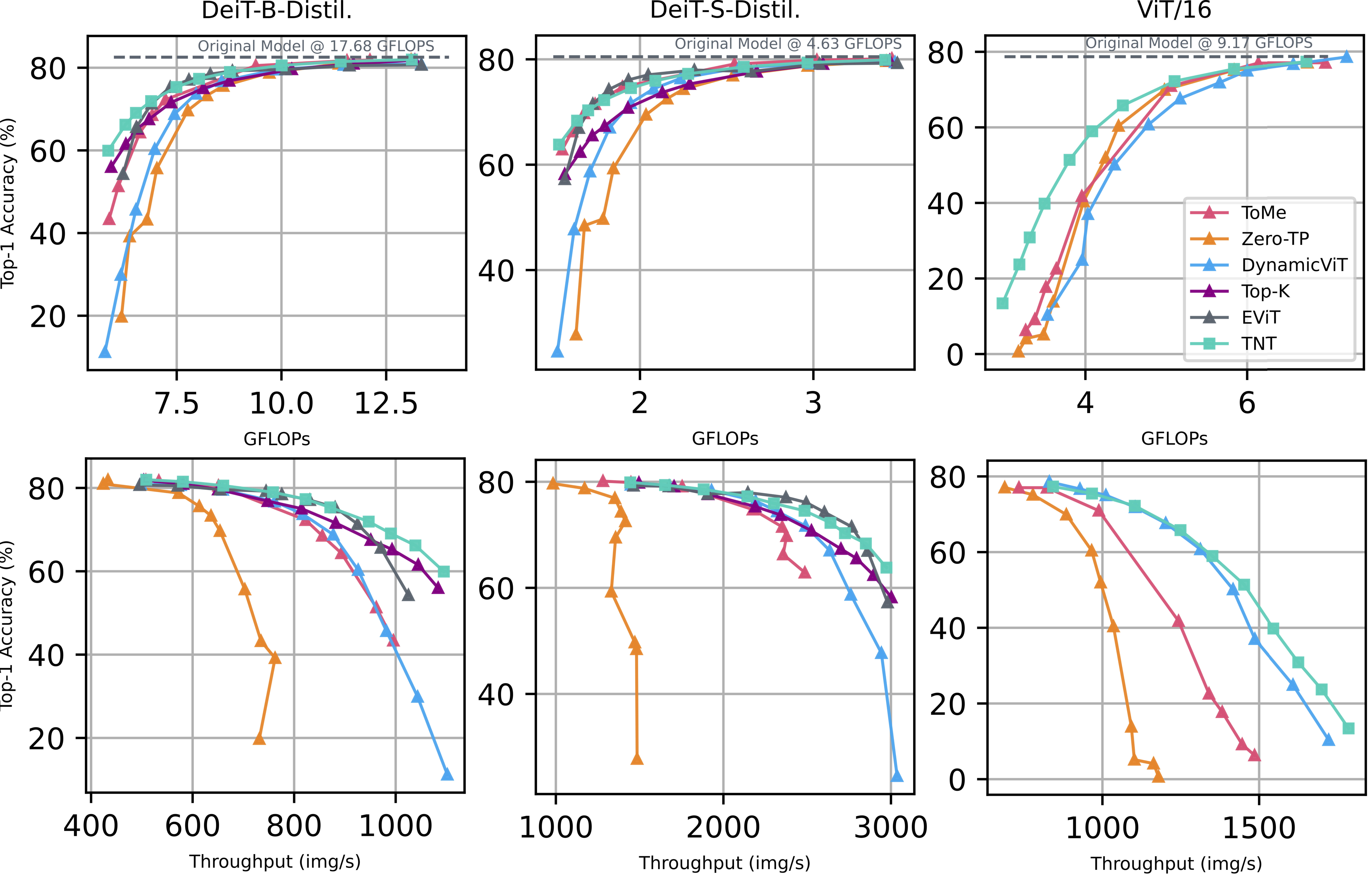}  
    \caption{\textbf{Multi-layer Pruning results:}
    We plot the Top-1 Accuracy in the ImageNet-1k validation set for each of the pruning methods as a function of computational efficiency, in the \textbf{top row} measured by GFLOPs and in the \textbf{bottom row} measured by throughput, for \textbf{multi-layer pruning}. The base model is DeiT-B-Distil in the \textbf{first column}, DeiT-S-Distil. in the \textbf{second column}, and ViT/16 in the \textbf{third column}. 
     Note that the mean-pooled token embedding ViT in the \textbf{third column} has no \texttt{CLS} token, and thus EViT and Top-K cannot be applied to it.
    }
    \label{fig:multi_layer}
\end{figure*}

\section{Experiments} \label{sec:experiments}

We conduct a series of experiments on pre-trained vision transformer models, including DeiT (Tiny\footnote{DeiT-Tiny results relegated to the Appendix.}, Small, Base)~\cite{touvron2021training} and ViT/16~\cite{dosovitskiy2020image}, each trained on ImageNet-1K \cite{deng2009imagenet}. We then compare our model with previously proposed token-dropping models, as well as two simple baseline schemes: DynamicViT \cite{rao2021dynamicvit}, ZeroTP\cite{wang2024zero}, ToMe \cite{bolya2022token}, Top-K \cite{haurum2023tokens}, EViT\cite{haurum2023tokens} and random dropping. Because ATS \cite{fayyaz2022adaptive} adaptively prunes a variable number of tokens per image, it does not allow a systematic sweep of the keep rate. Therefore, we do not include it as a baseline. We also validate our individual design choices in an ablation study (Section~\ref{subsec:ablation_study}). 

Haurum et al. \cite{haurum2023tokens} identifies \texttt{CLS} token attention based methods as state-of-the-art \cite{haurum2023tokens,fayyaz2022adaptive}, with a naive baseline (``Top-K'') nearly outperforming all others. These methods cannot be applied to models without CLS tokens, or where the CLS token is mismatched with the test-time task. However, our proposed method, which operates on similar principles but without the CLS token, can be applied in such settings, as demonstrated by experiments on the mean-pooled token embedding ViT.

We first show qualitative results for pruning $50\%$ of tokens at varying layers (single-layer pruning at layers 1-5) on the ImageNet-1K validation dataset in Figure ~\ref{fig:visualization}. As seen at left in the figure, extremely early layers result in relevant tokens being dropped, and irrelevant tokens being included; this matches with results from previous literature \cite{rao2021dynamicvit, fayyaz2022adaptive, wang2024zero}. Results further indicate that the the importance score based on noise signal term $\alpha$ becomes more reasonable in latter layers, performing well on a variety of qualitatively different images (tiny objects, mixed foreground and background, low contrast separation).
Additional examples are provided in the Appendix.


We conduct a comprehensive evaluation for each method by sweeping the token keep rate (K) from low to high-token regime, as detailed in Sections~\ref{subsec:single_layer_sweep} and~\ref{subsec:multiple_layers_sweep}. We experiment with both single-layer and multi-layer pruning schema. In the single-layer scheme all token pruning occurs at a single layer, usually early in the network. In comparison, the multi-layer scheme has pruning blocks spread across the architecture.
The multi-layer scheme is generally more effective, albeit with additional computational cost, but due to the large number of possible parameters for multi-layer pruning introduced by the multiple keep rates (with possibly varying predictive accuracy), may be hard to effectively tune for all methods. This makes comparisons in the multi-layer scheme inexact, which is why we include the single layer experiments.
We evaluate each model in terms of Top-1 Accuracy, FLOPs, and throughput. Our model achieves top performance across most experiments and remains competitive on others.

\textbf{Experimental Setup:} All training and evaluations are conducted on a single compute node equipped with 8 NVIDIA A40 GPUs. The experiments are performed on the ImageNet-1K dataset~\cite{deng2009imagenet}, with all images resized to a resolution of 224px.
We evaluate the proposed method using only frozen pre-trained backbone methods, training the Noise Allocator weights for 40 epochs. And we set $\beta$ (a hyperparameter controlling the magnitude of added noise) to 0.02 for all base models while training. For testing, a single GPU to measure the computational performance.

\subsection{Rate Sweep for Single Layer Pruning} \label{subsec:single_layer_sweep}

The experiments of previous studies have shown that most models~\cite{rao2021dynamicvit, fayyaz2022adaptive, wang2024zero} refrain from dropping tokens in initial Transformer layers, as these layers contain limited information relevant to token importance.
For our experiments, we select the second layer for ViT and third layer for DeiT variants as the standard layers for single-layer token pruning. 
We use the standard pre-trained DeiT variants for all experiments for all models.
For all ViT experiments for all models, we use a modified ViT/16 with smaller embedding dimension $mlp\_dim=emb\_dim=768$. 

Plots for pruning performance on DeiT-S-Dist., DeiT-B-Dist., ViT/16 across different pruning models are presented in Figure~\ref{fig:single_layer}, for both Top-1 Accuracy versus GFLOPs and Top-1 Accuracy versus throughput (images per second). In general throughput is not one-to-one with GFLOPs due to differing parallelism between models. Performance between all methods converge to the base model case as the keep rate approaches 1.0. For decreasing $K$ however, the proposed model shows superior performance until $K\leq0.25$, at which point baseline models overtake the proposed model, albeit with all models experiencing strong performance degradation. In general our results for the baseline models mirror that of Haurum et al. 2023 \cite{haurum2023tokens}, and similar to their report we also find that the Top-K baseline is generally the strongest where it can be applied (i.e., for models incorporating the \texttt{CLS} token), aside from our proposed method.

Overall, the number of tokens pruned is consistent across models for a given keep rate.
For TNT, we select top $(NK+s)$ at the Noise Allocator stage and prune $s$ tokens at the similarity pruning stage for a given keep rate $K$ and total number of tokens $N$. We set $s$ to be 25 and 30 for experiments involving DeiT and ViT, respectively.
We apply the same setting to Zero-TP~\cite{wang2024zero}, as it also utilizes similarity-based pruning.
We could not reproduce the results of Yin et al.~\cite{yin2022vit} in our own experimental context.
In Table~\ref{tab:single_layer}, we report specific numerical results for ViT/16 and DeiT-S-Distl, contingent on keep rates $K$, instead of GFLOPs or Throughput as in the plots. Complete numerical results and details such as keep rates for each experiment are also provided in the Appendix.
For ToMe~\cite{bolya2022token}, it is impossible to prune more than 50\% of tokens within a single layer, so lower single-layer pruning is not possible.

\subsection{Rate Sweep at Multi-layer Pruning} \label{subsec:multiple_layers_sweep}

We evaluate the performance of token pruning across multiple layers as this better reflects a performant pruning system, with the caveat that equal comparison conditions are more difficult to ensure. For the token keep rate at each layer and the specific layers chosen for token pruning (i.e. pruning locations), we strictly follow the instructions on the original paper for each model. For Zero-TP~\cite{wang2024zero}, tokens are pruned at layers [1, 3, 6, 9, 11], where layers [1, 11] perform similarity-based pruning only; for ToMe~\cite{bolya2022token}, pruning occurs at every layer; and for EViT \cite{liang2022not} and DynamicViT~\cite{rao2021dynamicvit}, pruning is applied at layers [3, 6, 9] and [4, 7, 10], respectively. As Top-K \cite{haurum2023tokens} lacks specific pruning instructions, we align its pruning layers with those of TNT. We gradually sweep the keep rate at pruning locations to generate Figure~\ref{fig:multi_layer} for DeiT-B-Distil., DeiT-S-Distil., and ViT/16. We also provide partial numerical results in Table~\ref{tab:multi-layer}. All base models are the same as those used in the single-layer pruning section. Additional details on experimental setups and specific details for each model are provided in the Appendix.

For TNT, we prefer pruning tokens at earlier layers, as it provides stronger performance, and as shown in Figure~\ref{fig:visualization}, layers [3, 4, 5] serve as effective locations for token pruning. Rather than performing the similarity pruning stage at every pruning layer as in Zero-TP, we find it is better to prune \( s \) tokens before tokens are sent to the ViT blocks. For all experiments, \( s \) is set to 40. Our results indicate that TNT consistently shows strong performance in varying token (computation) ranges. As the proportion of pruned tokens increases, the importance of the retained tokens becomes more critical. TNT shows a larger performance gap compared to other models in the low-token regime.

\subsection{Ablation Study} 
\label{subsec:ablation_study}
We conducted an ablation experiment to justify inclusion of the similarity-based pruning by random partition, as opposed to token merger or sequential partition. We find a slight boost to performance with similarity based pruning, with a very minor bias towards the random partition method. Previous works~\cite{bolya2022token, bolya2023token} explore token merging, which can be viewed as another form of pruning. For our method, this approach did not give an improvement over similarity pruning.





\begin{table}[h!]
    \centering
    \setlength{\tabcolsep}{3pt} 
    \renewcommand{\arraystretch}{1.0} 
    \begin{tabular}{lcc|c}
        \hline
        \rowcolor{white}
        \textbf{Method} & \textbf{Acc.} & \textbf{GFLOPs} & \textbf{TP (imgs/s)} \\
        \hline
        \textcolor{weakgray}{Deit-B-Distil.~\cite{touvron2021training}(SL)} & \textcolor{weakgray}{82.55} & \textcolor{weakgray}{17.68} & \textcolor{weakgray}{379} \\
        TNT w/o Sim. Pruning & 81.33 & 12.29  & 546 \\
        TNT (Seq. Part.) & 81.52 & 12.30 & 536 \\
        TNT (Token Merge) & 81.54 & 12.30 & 538 \\
        
        \rowcolor{lavender} TNT (Random Part.) & 81.57 & 12.30 & 542 \\

        \hline
        \textcolor{weakgray}{Deit-B-Distil.~\cite{touvron2021training}(ML)} & \textcolor{weakgray}{82.55} & \textcolor{weakgray}{17.68} & \textcolor{weakgray}{379} \\
        TNT w/o Sim. Pruning & 80.55 & 11.03 & 605 \\
        TNT (Token Merge) & 81.41 & 11.41 & 575 \\
        \rowcolor{lavender} TNT (Random Part.) & 81.50  & 11.41 & 581 \\
        \hline
    \end{tabular}
    \caption{Ablation study for TNT in DeiT-S-Distil. ``SL'' is single-layer pruning; ``ML'' is multi-layer pruning; \textbf{Acc.} is \textbf{Top-1 Accuracy}, \textbf{TP} is the \textbf{Throughput}, measured in images-per-second.}
    \label{tab:ablation}
\end{table}




\subsection{Limitations}
\label{sec:limitations}

While the proposed TNT method provides relatively improved performance over other methods, important gaps remain in methodology. In this method, redundant tokens are not removed during training; while this can be performed through merging as in ToMe \cite{bolya2022token}, as experimental results show it is difficult to do in a stable manner.
Further, synergistic information between tokens is not considered; 
this seems less relevant to ImageNet classification, but likely would be useful in more complicated label structures (e.g., hierarchical/multi-class structures). This is also perhaps a deeper problem than redundant tokens, as it has combinatorial complexity in the order of the interactions considered. While the Information Bottleneck technically addresses this, estimating models which include interaction information are generally intractable unless the variables are transformed (which makes them useless for Token Pruning).

Beyond this, for better comparison measurements multi-layer dropping should be tuned for each method. This is prohibitively expensive, but could provide some less stable but higher capacity methods with a boost in performance.

Finally, in a deployment setting, hardware constraints will dictate keep-rates, which could be optimized for in the base-models directly (i.e., we could optimize a ViT for a $50\%$ keep rate, or for a specific memory structure). This optimization was not done here, nor do we propose any candidate deployment device constraints, but nevertheless should be considered for best performance.

 \section{Conclusion} \label{sec:conclusion}
In this work, we introduced a novel token pruning method within the Information Bottleneck framework, aiming to optimize vision transformers by continuous optimization of token pruning. Our extensive evaluations on the ImageNet dataset using ViT and DeiT architectures demonstrate state-of-the-art performance in the accuracy-computation trade-off compared to existing methods. Specifically, our method excels in low-token retention rates, maintaining high accuracy while significantly reducing computational loads. These results underscore the potential impact of our method in improving the efficiency of deploying vision transformers, particularly in resource-constrained applications.

{
    \small
    \bibliographystyle{ieeenat_fullname}
    \bibliography{main}

\begin{thebibliography}{22}
\providecommand{\natexlab}[1]{#1}
\providecommand{\url}[1]{\texttt{#1}}
\expandafter\ifx\csname urlstyle\endcsname\relax
  \providecommand{\doi}[1]{doi: #1}\else
  \providecommand{\doi}{doi: \begingroup \urlstyle{rm}\Url}\fi

\bibitem[Alemi et~al.(2016)Alemi, Fischer, Dillon, and Murphy]{alemi2016deep}
Alexander~A Alemi, Ian Fischer, Joshua~V Dillon, and Kevin Murphy.
\newblock Deep variational information bottleneck.
\newblock \emph{arXiv preprint arXiv:1612.00410}, 2016.

\bibitem[Blahut(1972)]{blahut1972computation}
Richard Blahut.
\newblock Computation of channel capacity and rate-distortion functions.
\newblock \emph{IEEE transactions on Information Theory}, 18\penalty0
  (4):\penalty0 460--473, 1972.

\bibitem[Bolya and Hoffman(2023)]{bolya2023token}
Daniel Bolya and Judy Hoffman.
\newblock Token merging for fast stable diffusion.
\newblock In \emph{Proceedings of the IEEE/CVF conference on computer vision
  and pattern recognition}, pages 4599--4603, 2023.

\bibitem[Bolya et~al.(2022)Bolya, Fu, Dai, Zhang, Feichtenhofer, and
  Hoffman]{bolya2022token}
Daniel Bolya, Cheng-Yang Fu, Xiaoliang Dai, Peizhao Zhang, Christoph
  Feichtenhofer, and Judy Hoffman.
\newblock Token merging: Your vit but faster.
\newblock \emph{arXiv preprint arXiv:2210.09461}, 2022.

\bibitem[Cover(1999)]{cover1999elements}
Thomas~M Cover.
\newblock \emph{Elements of information theory}.
\newblock John Wiley \& Sons, 1999.

\bibitem[Deng et~al.(2009)Deng, Dong, Socher, Li, Li, and
  Fei-Fei]{deng2009imagenet}
Jia Deng, Wei Dong, Richard Socher, Li-Jia Li, Kai Li, and Li Fei-Fei.
\newblock Imagenet: A large-scale hierarchical image database.
\newblock In \emph{2009 IEEE conference on computer vision and pattern
  recognition}, pages 248--255. Ieee, 2009.

\bibitem[Dosovitskiy(2020)]{dosovitskiy2020image}
Alexey Dosovitskiy.
\newblock An image is worth 16x16 words: Transformers for image recognition at
  scale.
\newblock \emph{arXiv preprint arXiv:2010.11929}, 2020.

\bibitem[Fayyaz et~al.(2022)Fayyaz, Koohpayegani, Jafari, Sengupta, Joze,
  Sommerlade, Pirsiavash, and Gall]{fayyaz2022adaptive}
Mohsen Fayyaz, Soroush~Abbasi Koohpayegani, Farnoush~Rezaei Jafari, Sunando
  Sengupta, Hamid Reza~Vaezi Joze, Eric Sommerlade, Hamed Pirsiavash, and
  J{\"u}rgen Gall.
\newblock Adaptive token sampling for efficient vision transformers.
\newblock In \emph{European Conference on Computer Vision}, pages 396--414.
  Springer, 2022.

\bibitem[Goyal et~al.(2020)Goyal, Choudhury, Raje, Chakaravarthy, Sabharwal,
  and Verma]{goyal2020power}
Saurabh Goyal, Anamitra~Roy Choudhury, Saurabh Raje, Venkatesan Chakaravarthy,
  Yogish Sabharwal, and Ashish Verma.
\newblock Power-bert: Accelerating bert inference via progressive word-vector
  elimination.
\newblock In \emph{International Conference on Machine Learning}, pages
  3690--3699. PMLR, 2020.

\bibitem[Haurum et~al.(2023)Haurum, Escalera, Taylor, and
  Moeslund]{haurum2023tokens}
Joakim~Bruslund Haurum, Sergio Escalera, Graham~W Taylor, and Thomas~B
  Moeslund.
\newblock Which tokens to use? investigating token reduction in vision
  transformers.
\newblock In \emph{Proceedings of the IEEE/CVF International Conference on
  Computer Vision}, pages 773--783, 2023.

\bibitem[Kim and Cho(2020)]{kim2020length}
Gyuwan Kim and Kyunghyun Cho.
\newblock Length-adaptive transformer: Train once with length drop, use anytime
  with search.
\newblock \emph{arXiv preprint arXiv:2010.07003}, 2020.

\bibitem[Kim et~al.(2022)Kim, Shen, Thorsley, Gholami, Kwon, Hassoun, and
  Keutzer]{kim2022learned}
Sehoon Kim, Sheng Shen, David Thorsley, Amir Gholami, Woosuk Kwon, Joseph
  Hassoun, and Kurt Keutzer.
\newblock Learned token pruning for transformers.
\newblock In \emph{Proceedings of the 28th ACM SIGKDD Conference on Knowledge
  Discovery and Data Mining}, pages 784--794, 2022.

\bibitem[Kingma(2013)]{kingma2013auto}
Diederik~P Kingma.
\newblock Auto-encoding variational bayes.
\newblock \emph{arXiv preprint arXiv:1312.6114}, 2013.

\bibitem[Liang et~al.(2022)Liang, Ge, Tong, Song, Wang, and Xie]{liang2022not}
Youwei Liang, Chongjian Ge, Zhan Tong, Yibing Song, Jue Wang, and Pengtao Xie.
\newblock Not all patches are what you need: Expediting vision transformers via
  token reorganizations.
\newblock \emph{arXiv preprint arXiv:2202.07800}, 2022.

\bibitem[Page(1999)]{page1999pagerank}
Lawrence Page.
\newblock The pagerank citation ranking: Bringing order to the web.
\newblock Technical report, Technical Report, 1999.

\bibitem[Rao et~al.(2021)Rao, Zhao, Liu, Lu, Zhou, and
  Hsieh]{rao2021dynamicvit}
Yongming Rao, Wenliang Zhao, Benlin Liu, Jiwen Lu, Jie Zhou, and Cho-Jui Hsieh.
\newblock Dynamicvit: Efficient vision transformers with dynamic token
  sparsification.
\newblock \emph{Advances in neural information processing systems},
  34:\penalty0 13937--13949, 2021.

\bibitem[Tishby et~al.(2000)Tishby, Pereira, and Bialek]{tishby2000information}
Naftali Tishby, Fernando~C Pereira, and William Bialek.
\newblock The information bottleneck method.
\newblock \emph{arXiv preprint physics/0004057}, 2000.

\bibitem[Touvron et~al.(2021)Touvron, Cord, Douze, Massa, Sablayrolles, and
  J{\'e}gou]{touvron2021training}
Hugo Touvron, Matthieu Cord, Matthijs Douze, Francisco Massa, Alexandre
  Sablayrolles, and Herv{\'e} J{\'e}gou.
\newblock Training data-efficient image transformers \& distillation through
  attention.
\newblock In \emph{International conference on machine learning}, pages
  10347--10357. PMLR, 2021.

\bibitem[Vaswani(2017)]{vaswani2017attention}
A Vaswani.
\newblock Attention is all you need.
\newblock \emph{Advances in Neural Information Processing Systems}, 2017.

\bibitem[Wang et~al.(2024)Wang, Dedhia, and Jha]{wang2024zero}
Hongjie Wang, Bhishma Dedhia, and Niraj~K Jha.
\newblock Zero-tprune: Zero-shot token pruning through leveraging of the
  attention graph in pre-trained transformers.
\newblock In \emph{Proceedings of the IEEE/CVF Conference on Computer Vision
  and Pattern Recognition}, pages 16070--16079, 2024.

\bibitem[Xu et~al.(2022)Xu, Zhang, Zhang, Sheng, Li, Dong, Zhang, Xu, and
  Sun]{xu2022evo}
Yifan Xu, Zhijie Zhang, Mengdan Zhang, Kekai Sheng, Ke Li, Weiming Dong, Liqing
  Zhang, Changsheng Xu, and Xing Sun.
\newblock Evo-vit: Slow-fast token evolution for dynamic vision transformer.
\newblock In \emph{Proceedings of the AAAI Conference on Artificial
  Intelligence}, pages 2964--2972, 2022.

\bibitem[Yin et~al.(2022)Yin, Vahdat, Alvarez, Mallya, Kautz, and
  Molchanov]{yin2022vit}
Hongxu Yin, Arash Vahdat, Jose~M Alvarez, Arun Mallya, Jan Kautz, and Pavlo
  Molchanov.
\newblock A-vit: Adaptive tokens for efficient vision transformer.
\newblock In \emph{Proceedings of the IEEE/CVF conference on computer vision
  and pattern recognition}, pages 10809--10818, 2022.

\end{thebibliography}
}

\clearpage
\setcounter{page}{1}
\maketitlesupplementary

\renewcommand\thesection{\Alph{section}}
\setcounter{section}{0}

\section{Implementation Code}
Figure~\ref{fig:code} is an implementation of our “VisionTransformerWithTNT” in PyTorch.

\begin{figure*}[h!]
\centering
\lstset{style=custom_python}
\begin{lstlisting}
class VisionTransformerWithTNT(VisionTransformer):
    def __init__(self, *args, **kwargs):
        super().__init__(*args, **kwargs)
        # Parameters introduced: Add alpha heads to produce noise signal term
        @self.alpha_norm = kwargs['norm_layer'](self.embed_dim)
        self.alpha_heads = nn.ModuleList([
            nn.Linear(self.embed_dim, 1) for _ in range(kwargs['depth'])
        ])
        self.alpha_heads.apply(self._init_weights)@

    def forward_features(self, x):
        B = x.shape[0]
        x = self.patch_embed(x)

        cls_tokens = self.cls_token.expand(B, -1, -1)
        x = torch.cat((cls_tokens, x), dim=1)
        x = x + self.pos_embed
        x = self.pos_drop(x)
        for i, (blk, alpha_head) in enumerate(zip(self.blocks, self.alpha_heads)):
            x = blk(x)
            # Noise allocator: To add noise to token embeddings at 1-5 layers while fine-tuning
            @if self.training and i < 5:
                alpha = alpha_head(x[:, 1:])
                alpha = torch.softmax(alpha.squeeze(-1), dim=-1)
                alpha = 1 - alpha
                noise = torch.randn_like(x[:, 1:]) * alpha.unsqueeze(-1).repeat(1,1,x.size(-1))
                zero_noise = torch.zeros_like(cls_tokens)
                noise = torch.cat((zero_noise, noise), dim=1)
                x = self.alpha_norm(x)
                x = x + 0.02 * noise@
        x = self.norm(x)
        return x[:, 0]

    def forward(self, x):
        x = self.forward_features(x)
        x = self.head(x)
        return x
\end{lstlisting}
\caption{Python implementation of \textbf{VisionTransformerWithTNT} class. Codes highlighted with \textcolor{brown}{brown} are the main modifications. \textbf{VisionTransformer} class is taken from \url{https://github.com/rwightman/pytorch-image-models/blob/master/timm/models/vision_transformer.py}. We make simple modifications to allocate noise to token embeddings while fine-tuning.}
\label{fig:code}
\end{figure*}

\section{Additional Experiment Details}
Full results for plots and tables in the main paper including DeiT-Tiny-Distil.~\cite{touvron2021training}. For all results, \textbf{Acc.} is Top-1
Accuracy, \textbf{TP} is the Throughput, measured in images-per-second.

\textbf{Single-layer pruning settings:} \textbf{K} is the keep rate. We divide K values into two tables: K = 0.8 to 0.45 represent the high-token regime, while K = 0.4 to 0.2 correspond to the low-token regime. Highlighted rows achieve the highest accuracy in a specific token regime. For Top-K~\cite{liang2022not}, GFLOPs is relatively smaller as pruning occurs in the middle of the Transformer block. For clarity in visualizations of the computation-accuracy trade-off, data of Top-K corresponding to K=0.2 is omitted from the plots but is reported in the table.

\textbf{Multi-layer pruning settings:} For TNT, Top-K~\cite{liang2022not}, and Zero-TP~\cite{wang2024zero}, \textbf{Loc} and \textbf{Rate} denote the pruning locations (layers) and the pruning rate at each layer, respectively. \textbf{itr} is an additional parameter for Zero-TP only, which specifies the number of iterations for the PageRank algorithm at each pruning layer. For ToMe~\cite{bolya2022token}, pruning is performed at every layer, with $\bm{r}$ representing the number of tokens pruned per layer. To ensure GFLOPs alignment for better comparison, we only collect 8 data points per base model for ToMe (10 for other models). For DynamicViT~\cite{rao2021dynamicvit}, the keep rates for the three pruning layers are specified as [$\bm{\rho}$, $\bm{\rho}^2$, $\bm{\rho}^3$]. \textbf{Rate}, $\bm{r}$, and $\bm{\rho}$ vary across different experiments.

 Table~\ref{tab:fixed_param} provides an overview of the fixed hyperparameters used consistently in all experiments. For different base models, we conduct the hyperexperiments using the same set of parameters.

\begin{table}[h!]
    \centering
    \setlength{\tabcolsep}{3pt} 
    \renewcommand{\arraystretch}{1.0} 
    \begin{tabular}{l|cc}
        \hline
        \rowcolor{white}
        \textbf{Method} & \textbf{Param.}   \\
        \hline
        Top-K~\cite{liang2022not} & loc=[3, 4, 5] \\
        \hline
        \multirow{3}{*}{Zero-TP~\cite{wang2024zero}}
        & itr(single-layer)=50 \\
        & loc=[1, 3, 6, 9, 11]  \\
        & itr(multi-layer)=[30, 5, 5, 1, 1] \\
        \hline
        DynamicViT~\cite{rao2021dynamicvit} & loc=[3, 6, 9] \\
        \hline
        ToMe~\cite{bolya2022token} & loc[1:12]  \\
        \hline
        EViT~\cite{liang2022not} & loc[4, 7, 10]  \\
        \hline
        TNT (ours) & loc=[3, 4, 5]  \\

        \hline
    \end{tabular}
    \caption{overview of the fixed hyperparameters used consistently in all experiments.}
    \label{tab:fixed_param}
\end{table}

\subsection{Base Model: DeiT-Base-Distil.}
For single-layer pruning, high-token regime results are listed in Table~\ref{tab:deit_b_single_layer_high}; low-token regime results are listed in Table~\ref{tab:deit_b_single_layer_low}. For multi-layer pruning, high-token regime results are listed in Table~\ref{tab:deit_b_multi_layer_high}; low-token regime results are listed in Table~\ref{tab:deit_b_multi_layer_low}.

\begin{table}[h!]
    \centering
    \setlength{\tabcolsep}{3pt} 
    \renewcommand{\arraystretch}{1.0} 
    \begin{tabular}{c|lccc}
        \hline
        \rowcolor{white}
        & \textbf{Method} & \textbf{Acc.} & \textbf{GFLOPs} & \textbf{TP (imgs/s)} \\
        \hline
        & \textcolor{weakgray}{Deit-B-Distil.} & \textcolor{weakgray}{82.55} & \textcolor{weakgray}{17.68} & \textcolor{weakgray}{379} \\
        \hline
        \multirow{6}{*}{\rotatebox[origin=c]{90}{K = 0.8}}
        & Random Drop & 81.59 & 14.93 & - \\
        & Top-K~\cite{liang2022not} & 82.16 & 14.74 & 462 \\
        & Zero-TP~\cite{wang2024zero} & 82.25 & 14.95 & 411 \\
        & DynamicViT~\cite{rao2021dynamicvit} & 80.87 & 15.33 & 438 \\
        & ToMe~\cite{bolya2022token} & 81.86 & 14.74 & 447 \\
        & EViT~\cite{liang2022not} & 81.65 & 15.08 & 448 \\
        & \cellcolor{lavender}TNT (ours) & \cellcolor{lavender}82.31 & \cellcolor{lavender}14.93 & \cellcolor{lavender}455 \\
        \hline
        
        \multirow{6}{*}{\rotatebox[origin=c]{90}{K = 0.7}}
        & Random Drop & 80.79 & 13.64 & - \\
        & Top-K~\cite{liang2022not} & 81.63 & 13.36 & 507 \\
        & Zero-TP~\cite{wang2024zero} & 81.74 & 13.67 & 448 \\
        & DynamicViT~\cite{rao2021dynamicvit} & 80.66 & 14.00 & 481 \\
        & ToMe~\cite{bolya2022token} & 80.80 & 13.36 & 493 \\
        & EViT~\cite{liang2022not} & 81.49 & 13.85 & 486 \\
        & \cellcolor{lavender}TNT (ours) & \cellcolor{lavender}82.05 & \cellcolor{lavender}13.64 & \cellcolor{lavender}495 \\
        \hline
        
        \multirow{6}{*}{\rotatebox[origin=c]{90}{K = 0.6}}
        & Random Drop & 79.59 & 12.29 & - \\
        & Top-K~\cite{liang2022not} & 80.70 & 11.92 & 558 \\
        & Zero-TP~\cite{wang2024zero} & 80.58 & 12.32 & 488 \\
        & DynamicViT~\cite{rao2021dynamicvit} & 80.02 & 12.61 & 526 \\
        & ToMe~\cite{bolya2022token} & 78.28 & 11.92 & 546 \\
        & EViT~\cite{liang2022not} & 80.08 & 12.55 & 532 \\
        & \cellcolor{lavender}TNT (ours) & \cellcolor{lavender}81.57 & \cellcolor{lavender}12.30 & \cellcolor{lavender}542 \\
        \hline
        
        \multirow{6}{*}{\rotatebox[origin=c]{90}{K = 0.5}}
        & Random Drop & 77.72 & 11.02 & - \\
        & Top-K~\cite{liang2022not} & 79.25 & 10.56 & 635 \\
        & Zero-TP~\cite{wang2024zero} & 78.63 & 11.05 & 540 \\
        & DynamicViT~\cite{rao2021dynamicvit} & 79.06 & 11.31 & 588 \\
        & ToMe~\cite{bolya2022token} & 71.57 & 10.56 & 621 \\
        & EViT~\cite{liang2022not} & 79.48 & 11.27 & 589 \\
        & \cellcolor{lavender}TNT (ours) & \cellcolor{lavender}80.62 & \cellcolor{lavender}11.03 & \cellcolor{lavender}605 \\
        \hline
        
        \multirow{5}{*}{\rotatebox[origin=c]{90}{K = 0.45}}
        & Random Drop & 76.23 & 10.35 & - \\
        & Top-K~\cite{liang2022not} & 78.06 & 9.85 & 677 \\
        & Zero-TP~\cite{wang2024zero} & 77.15 & 10.38 & 571 \\
        & DynamicViT~\cite{rao2021dynamicvit} & 78.26 & 10.63 & 624 \\
        & EViT~\cite{liang2022not} & 79.1 & 10.70 & 619 \\
        & \cellcolor{lavender}TNT (ours) & \cellcolor{lavender}79.85 & \cellcolor{lavender}10.31 & \cellcolor{lavender}640 \\
        \hline
    \end{tabular}

    \caption{
    \textbf{Single-layer pruning} for DeiT-B-Distil. in high-token regime.
    }
    \label{tab:deit_b_single_layer_high}
\end{table}

\begin{table}[h!]
    \centering
    \setlength{\tabcolsep}{3pt} 
    \renewcommand{\arraystretch}{1.0} 
    \begin{tabular}{c|lccc}
        \hline
        \rowcolor{white}
        & \textbf{Method} & \textbf{Acc.} & \textbf{GFLOPs} & \textbf{TP (imgs/s)} \\
        \hline
        & \textcolor{weakgray}{Deit-B-Distil.} & \textcolor{weakgray}{82.55} & \textcolor{weakgray}{17.68} & \textcolor{weakgray}{379} \\
        \hline
        \multirow{5}{*}{\rotatebox[origin=c]{90}{K = 0.4}}
        & Random Drop & 74.08 & 9.70 & - \\
        & Top-K~\cite{liang2022not} & 76.53 & 9.14 & 728 \\
        & Zero-TP~\cite{wang2024zero} & 75.15 & 9.72 & 603 \\
        & DynamicViT~\cite{rao2021dynamicvit} & 77.01 & 9.95 & 668 \\
        & EViT~\cite{liang2022not} & 78.33 & 10.06 & 658 \\
        & \cellcolor{lavender}TNT (ours) & \cellcolor{lavender}78.89 & \cellcolor{lavender}9.70 & \cellcolor{lavender}684 \\
        \hline
        
        \multirow{5}{*}{\rotatebox[origin=c]{90}{K = 0.35}}
        & Random Drop & 70.57 & 9.04 & - \\
        & Top-K~\cite{liang2022not} & 74.26 & 8.43 & 786 \\
        & Zero-TP~\cite{wang2024zero} & 72.32 & 9.06 & 649 \\
        & DynamicViT~\cite{rao2021dynamicvit} & 75.28 & 9.27 & 710 \\
        & \cellcolor{lavender}EViT~\cite{liang2022not} & \cellcolor{lavender}77.31 & \cellcolor{lavender}9.43 & \cellcolor{lavender}698 \\
        & TNT (ours) & 77.22 & 9.04 & 723 \\
        \hline
        
        \multirow{5}{*}{\rotatebox[origin=c]{90}{K = 0.3}}
        & Random Drop & 64.89 & 8.38 & - \\
        & Top-K~\cite{liang2022not} & 70.68 & 7.73 & 853 \\
        & Zero-TP~\cite{wang2024zero} & 68.68 & 8.41 & 680 \\
        & DynamicViT~\cite{rao2021dynamicvit} & 72.69 & 8.59 & 770 \\
        & \cellcolor{lavender}EViT~\cite{liang2022not} & \cellcolor{lavender}75.87 & \cellcolor{lavender}8.80 & \cellcolor{lavender}749 \\
        & TNT (ours) & 74.75 & 8.38 & 786 \\
        \hline
        
        \multirow{5}{*}{\rotatebox[origin=c]{90}{K = 0.25}}
        & Random Drop & 57.12 & 7.80 & - \\
        & Top-K~\cite{liang2022not} & 65.95 & 7.10 & 925 \\
        & Zero-TP~\cite{wang2024zero} & 63.62 & 7.82 & 717 \\
        & DynamicViT~\cite{rao2021dynamicvit} & 68.87 & 7.99 & 823 \\
        & \cellcolor{lavender}EViT~\cite{liang2022not} & \cellcolor{lavender}73.54 & \cellcolor{lavender}8.17 & \cellcolor{lavender}801 \\
        & TNT (ours) & 70.76 & 7.80 & 834 \\
        \hline
        
        \multirow{5}{*}{\rotatebox[origin=c]{90}{K = 0.2}}
        & Random Drop & 44.17 & 7.14 & - \\
        & Top-K~\cite{liang2022not} & 57.51 & 6.40 & 1016 \\
        & Zero-TP~\cite{wang2024zero} & 55.28 & 7.17 & 779 \\
        & DynamicViT~\cite{rao2021dynamicvit} & 61.86 & 7.32 & 904 \\
        & \cellcolor{lavender}EViT~\cite{liang2022not} & \cellcolor{lavender}70.4 & \cellcolor{lavender}7.61 & \cellcolor{lavender}856 \\
        & TNT (ours) & 62.62 & 7.15 & 907 \\
        \hline
    \end{tabular}

    \caption{
    \textbf{Single-layer pruning} for DeiT-B-Distil. in low-token regime.
    }
    \label{tab:deit_b_single_layer_low}
\end{table}

\begin{table*}[h!]
    \centering
    \setlength{\tabcolsep}{3pt} 
    \renewcommand{\arraystretch}{1.0} 
        \begin{tabular}{c|lcccc}
        \hline
        \rowcolor{white}
         & \textbf{Method} & \textbf{Acc.} & \textbf{GFLOPs} & \textbf{TP (imgs/s)} & \textbf{Param.}\\
        \hline
        & \textcolor{weakgray}{Deit-B-Distil.} & \textcolor{weakgray}{82.55} & \textcolor{weakgray}{17.68} & \textcolor{weakgray}{379} & \textcolor{weakgray}{-}\\
        \hline
        \multirow{5}{*}{\rotatebox[origin=c]{90}{$\tiny\text{GFLOPs } \text{\tiny $ \approx 13.0$}$}}
        & Top-K & 81.82 & 13.18 & 503 & Rate=[.9, .9, .8] \\ 
        & Zero-TP~\cite{wang2024zero} & 81.92 & 13.10 & 433 & Rate=[1., .9, .9, .9, 1.]\\
        & DynamicViT~\cite{rao2021dynamicvit} & 80.91 & 13.34 & 498 & $\rho$=.8\\
        & ToMe~\cite{bolya2022token} & 81.86 & 12.11 & 508 & $r$=8\\
        & EViT~\cite{liang2022not} & 80.70 & 13.34 & 496 & $\rho$=.8\\
        & \cellcolor{lavender}TNT (ours) & \cellcolor{lavender}81.97 & \cellcolor{lavender}13.11 & \cellcolor{lavender}509 &
        \cellcolor{lavender}Rate=[1., .95, .95] \\
        \hline

        \multirow{5}{*}{\rotatebox[origin=c]{90}{$\tiny\text{GFLOPs } \text{\tiny $ \approx 11.5$}$}}
        & Top-K & 80.96 & 11.72 & 581 & Rate=[.85, .8, .8] \\ 
        & Zero-TP~\cite{wang2024zero} & 80.97 & 11.37 & 424 & Rate=[1., .8, .8, .9, 1.]\\
        & DynamicViT~\cite{rao2021dynamicvit} & 80.67 & 11.49 & 576 & $\rho$=.7\\
        & \cellcolor{lavender}ToMe~\cite{bolya2022token} & \cellcolor{lavender}81.71 & \cellcolor{lavender}11.56 & \cellcolor{lavender}533 & \cellcolor{lavender}$r$=11 \\
        & EViT~\cite{liang2022not} & 80.52 & 11.62 & 571 & $\rho$=.7\\
        & TNT (ours) & 81.50 & 11.41 & 581 & Rate=[.9, .9, .9]\\
        \hline

        \multirow{5}{*}{\rotatebox[origin=c]{90}{$\tiny\text{GFLOPs } \text{\tiny $ \approx 10.0$}$}}
        & Top-K & 79.63 & 10.25 & 650 & Rate=[.85, .7, .7] \\ 
        & Zero-TP~\cite{wang2024zero} & 78.77 & 9.71 & 573 & Rate=[1., .7, .7, .8, 1.]\\
        & DynamicViT~\cite{rao2021dynamicvit} & 79.55 & 9.88 & 659 & $\rho$=.6\\
        & ToMe~\cite{bolya2022token} & 80.51 & 9.39 & 651 & $r$=15\\
        & EViT~\cite{liang2022not} & 79.60 & 10.10 & 654 & $\rho$=.6\\
        & \cellcolor{lavender}TNT (ours) & \cellcolor{lavender}80.56 & \cellcolor{lavender}10.01 & \cellcolor{lavender}660 &
        \cellcolor{lavender}Rate=[.85, .85, .8]\\
        \hline

        \multirow{4}{*}{\rotatebox[origin=c]{90}{$\tiny\text{GFLOPs } \text{\tiny $ \approx 8.5$}$}}
        & Top-K & 76.84 & 8.75 & 748 & Rate=[.7, .7, .65] \\ 
        & Zero-TP~\cite{wang2024zero} & 75.60 & 8.61 & 613 & Rate=[1., .6, .7, .7, 1.]\\
        & DynamicViT~\cite{rao2021dynamicvit} & 76.73 & 8.57 & 761 & $\rho$=.5\\
        & EViT~\cite{liang2022not} & 79.24 & 8.83 & 745 & $\rho$=.5\\
        & \cellcolor{lavender}TNT (ours) & \cellcolor{lavender}78.96 & \cellcolor{lavender}8.77 & \cellcolor{lavender}758 & \cellcolor{lavender}Rate=[.75, .85, .7]\\
        \hline

        \multirow{4}{*}{\rotatebox[origin=c]{90}{$\tiny\text{GFLOPs } \text{\tiny $ \approx 8.0$}$}}
        & Top-K & 75.04 & 8.14 & 814 & Rate=[.65, .65, .65] \\ 
        & Zero-TP~\cite{wang2024zero} & 73.32 & 8.22 & 636 & Rate=[1., .6, .6, .7, 1.]\\
        & DynamicViT~\cite{rao2021dynamicvit} & 73.71 & 7.96 & 817 & $\rho$=.45\\
        & \cellcolor{lavender}EViT~\cite{liang2022not} & \cellcolor{lavender}78.46 & \cellcolor{lavender}8.30 & \cellcolor{lavender}776 & \cellcolor{lavender}$\rho$=.45\\
        & TNT (ours) & 77.31 & 8.03 & 822 & Rate=[.65, .85, .7]\\
        \hline
    \end{tabular}

    \caption{
    \textbf{Multi-layer pruning} for DeiT-B-Distil. in high-token regime.
    }
    \label{tab:deit_b_multi_layer_high}
\end{table*}

\begin{table*}[h!]
    \centering
    \setlength{\tabcolsep}{3pt} 
    \renewcommand{\arraystretch}{1.0} 
    \begin{tabular}{c|lcccc}
        \hline
        \rowcolor{white}
        & \textbf{Method} & \textbf{Acc.} & \textbf{GFLOPs} & \textbf{TP (imgs/s)} & \textbf{Param.} \\
        \hline
        & \textcolor{weakgray}{Deit-B-Distil.} & \textcolor{weakgray}{82.55} & \textcolor{weakgray}{17.68} & \textcolor{weakgray}{379} & \textcolor{weakgray}{-}\\
        \hline
        \multirow{5}{*}{\rotatebox[origin=c]{90}{$\tiny\text{GFLOPs } \text{\tiny $ \approx 7.5$}$}}
        & Top-K & 71.59 & 7.37 & 881 & Rate=[.6, .6, .6] \\ 
        & Zero-TP~\cite{wang2024zero} & 69.66 & 7.76 & 654 & Rate=[1., .5, .5, .6, 1.] \\
        & DynamicViT~\cite{rao2021dynamicvit} & 68.79 & 7.45 & 877 & $\rho$=.4 \\
        & ToMe~\cite{bolya2022token} & 72.29 & 7.23 & 823 & $r$=20\\
        & \cellcolor{lavender}EViT~\cite{liang2022not} & \cellcolor{lavender}77.09 & \cellcolor{lavender}7.79 & \cellcolor{lavender}831 & \cellcolor{lavender}$\rho$=.4\\
        & TNT (ours) & 75.33 & 7.49 & 871 & Rate=[.6, .8, .7] \\
        \hline

        \multirow{5}{*}{\rotatebox[origin=c]{90}{$\tiny\text{GFLOPs } \text{\tiny $ \approx 7.0$}$}}
        & Top-K & 67.51 & 6.84 & 948 & Rate=[.6, .5, .6] \\ 
        & Zero-TP~\cite{wang2024zero} & 55.65 & 7.03 & 703 & Rate=[1., .4, .4, .4, 1.] \\
        & DynamicViT~\cite{rao2021dynamicvit} & 60.30 & 6.97 & 926 & $\rho$=.35 \\
        & ToMe~\cite{bolya2022token} & 68.53 & 6.91 & 855 & $r$=21\\
        & \cellcolor{lavender}EViT~\cite{liang2022not} & \cellcolor{lavender}75.32 & \cellcolor{lavender}7.34 & \cellcolor{lavender}881 & \cellcolor{lavender}$\rho$=.35\\
        & TNT (ours) & 71.92 & 6.89 & 947 & Rate=[.5, .8, .7] \\
        \hline

        \multirow{5}{*}{\rotatebox[origin=c]{90}{$\tiny\text{GFLOPs } \text{\tiny $ \approx 6.6$}$}}
        & Top-K & 65.24 & 6.56 & 993 & Rate=[.55, .5, .6] \\ 
        & Zero-TP~\cite{wang2024zero} & 43.24 & 6.79 & 735 & Rate=[1., .4, .3, .4, 1.] \\
        & DynamicViT~\cite{rao2021dynamicvit} & 45.65 & 6.53 & 982 & $\rho$=.3 \\ 
        & ToMe~\cite{bolya2022token} & 64.33 & 6.62 & 893 & $r$=22 \\ 
        & \cellcolor{lavender}EViT~\cite{liang2022not} & \cellcolor{lavender}71.32 & \cellcolor{lavender}6.90 & \cellcolor{lavender}925 & \cellcolor{lavender}$\rho$=.3\\
        & TNT (ours) & 69.06 & 6.53 & 991 & Rate=[.5, .7, .7] \\
        \hline

        \multirow{5}{*}{\rotatebox[origin=c]{90}{$\tiny\text{GFLOPs } \text{\tiny $ \approx 6.2$}$}}
        & Top-K & 61.54 & 6.29 & 1044 & Rate=[.55, .5, .5] \\ 
        & Zero-TP~\cite{wang2024zero} & 39.17 & 6.37 & 762 & Rate=[1., .3, .4, .4, 1.] \\
        & DynamicViT~\cite{rao2021dynamicvit} & 29.87 & 6.17 & 1043 & $\rho$=.25 \\
        & ToMe~\cite{bolya2022token} & 51.32 & 6.11 & 962 & $r$=24 \\
        & EViT~\cite{liang2022not} & 65.62 & 6.54 & 971 & $\rho$=.25\\
        & \cellcolor{lavender}TNT (ours) & \cellcolor{lavender}66.21 & \cellcolor{lavender}6.28 & \cellcolor{lavender}1039 & \cellcolor{lavender}Rate=[.5, .7, .6] \\
        \hline

        \multirow{5}{*}{\rotatebox[origin=c]{90}{$\tiny\text{GFLOPs } \text{\tiny $ \approx 5.7$}$}}
        & Top-K & 55.98 & 5.93 & 1084 & Rate=[.5, .45, .5] \\ 
        & Zero-TP~\cite{wang2024zero} & 19.76 & 6.18 & 731 & Rate=[1., .3, .3, .4, 1.] \\
        & DynamicViT~\cite{rao2021dynamicvit} & 11.20 & 5.79 & 1102 & $\rho$=.2 \\
        & ToMe~\cite{bolya2022token} & 43.37 & 5.89 & 995 & $r$=25 \\
        & EViT~\cite{liang2022not} & 54.27 & 6.22 & 1025 & $\rho$=.2\\
        & \cellcolor{lavender}TNT (ours) & \cellcolor{lavender}59.93 & \cellcolor{lavender}5.87 & \cellcolor{lavender}1095 & \cellcolor{lavender}Rate=[.5, .6, .55] \\
        \hline
    \end{tabular}

    \caption{
    \textbf{Multi-layer pruning} for DeiT-B-Distil. in low-token regime.
    }
    \label{tab:deit_b_multi_layer_low}
\end{table*}

\subsection{Base Model: DeiT-Small-Distil.}

For single-layer pruning, high-token regime results are listed in Table~\ref{tab:deit_s_single_layer_high}; low-token regime results are listed in Table~\ref{tab:deit_s_single_layer_low}. For multi-layer pruning, high-token regime results are listed in Table~\ref{tab:deit_s_multi_layer_high}; low-token regime results are listed in Table~\ref{tab:deit_s_multi_layer_low}.

\begin{table}[h!]
    \centering
    \setlength{\tabcolsep}{3pt} 
    \renewcommand{\arraystretch}{1.0} 
    \begin{tabular}{c|lccc}
        \hline
        \rowcolor{white}
        & \textbf{Method} & \textbf{Acc.} & \textbf{GFLOPs} & \textbf{TP (imgs/s)} \\
        \hline
        & \textcolor{weakgray}{Deit-S-Distil.} & \textcolor{weakgray}{80.49} & \textcolor{weakgray}{4.63} & \textcolor{weakgray}{1150} \\
        \hline
        \multirow{6}{*}{\rotatebox[origin=c]{90}{K = 0.8}}
        & Random Drop & 79.54 & 3.90 & - \\
        & \cellcolor{lavender}Top-K~\cite{liang2022not} & \cellcolor{lavender}80.12 & \cellcolor{lavender}3.85 & \cellcolor{lavender}1338 \\
        & Zero-TP~\cite{wang2024zero} & 79.99 & 3.91 & 1156 \\
        & DynamicViT~\cite{rao2021dynamicvit} & 79.39 & 3.99 & 1255 \\
        & ToMe~\cite{bolya2022token} & 80.07 & 3.85 & 1290 \\
        & EViT~\cite{liang2022not} & 78.77 & 3.86 & 1337\\
        & TNT (ours) & 80.11 & 3.90 & 1316 \\
        \hline

        \multirow{6}{*}{\rotatebox[origin=c]{90}{K = 0.7}}
        & Random Drop & 78.91 & 3.55 & - \\
        & Top-K~\cite{liang2022not} & 79.69 & 3.48 & 1478 \\
        & Zero-TP~\cite{wang2024zero} & 79.55 & 3.57 & 1243 \\
        & DynamicViT~\cite{rao2021dynamicvit} & 79.25 & 3.64 & 1381 \\
        & ToMe~\cite{bolya2022token} & 79.24 & 3.49 & 1415 \\
        & EViT~\cite{liang2022not} & 78.44 & 3.50 & 1472\\
        & \cellcolor{lavender}TNT (ours) & \cellcolor{lavender}79.82 & \cellcolor{lavender}3.56 & \cellcolor{lavender}1427 \\
        \hline

        \multirow{6}{*}{\rotatebox[origin=c]{90}{K = 0.6}}
        & Random Drop & 77.83 & 3.20 & - \\
        & Top-K~\cite{liang2022not} & 78.89 & 3.11 & 1655 \\
        & Zero-TP~\cite{wang2024zero} & 78.94 & 3.21 & 1343 \\
        & DynamicViT~\cite{rao2021dynamicvit} & 78.66 & 3.28 & 1528 \\
        & ToMe~\cite{bolya2022token} & 77.40 & 3.11 & 1596 \\
        & EViT~\cite{liang2022not} & 76.07 & 3.12 & 1652\\
        & \cellcolor{lavender}TNT (ours) & \cellcolor{lavender}79.29 & \cellcolor{lavender}3.20 & \cellcolor{lavender}1587 \\
        \hline

        \multirow{6}{*}{\rotatebox[origin=c]{90}{K = 0.5}}
        & Random Drop & 76.57 & 2.87 & - \\
        & Top-K~\cite{liang2022not} & 77.69 & 2.75 & 1854 \\
        & Zero-TP~\cite{wang2024zero} & 77.74 & 2.88 & 1463 \\
        & DynamicViT~\cite{rao2021dynamicvit} & 77.96 & 2.94 & 1706 \\
        & ToMe~\cite{bolya2022token} & 73.13 & 2.75 & 1791 \\
        & EViT~\cite{liang2022not} & 76.28 & 2.75 & 1849\\
        & \cellcolor{lavender}TNT (ours) & \cellcolor{lavender}78.65 & \cellcolor{lavender}2.87 & \cellcolor{lavender}1768 \\
        \hline

        \multirow{5}{*}{\rotatebox[origin=c]{90}{K = 0.45}}
        & Random Drop & 75.61 & 2.69 & - \\
        & Top-K~\cite{liang2022not} & 76.70 & 2.57 & 1995 \\
        & Zero-TP~\cite{wang2024zero} & 76.97 & 2.71 & 1514 \\
        & DynamicViT~\cite{rao2021dynamicvit} & 77.22 & 2.76 & 1800 \\
        & EViT~\cite{liang2022not} & 75.65 & 2.58 & 1962\\
        & \cellcolor{lavender}TNT (ours) & \cellcolor{lavender}78.06 & \cellcolor{lavender}2.70 & \cellcolor{lavender}1878 \\
        \hline
    \end{tabular}

    \caption{
    \textbf{Single-layer pruning} for DeiT-S-Distil. in high-token regime.
    }
    \label{tab:deit_s_single_layer_high}
\end{table}

\begin{table}[h!]
    \centering
    \setlength{\tabcolsep}{3pt} 
    \renewcommand{\arraystretch}{1.0} 
    \begin{tabular}{c|lccc}
        \hline
        \rowcolor{white}
        & \textbf{Method} & \textbf{Acc.} & \textbf{GFLOPs} & \textbf{TP (imgs/s)} \\
        \hline
        & \textcolor{weakgray}{Deit-S-Distil.} & \textcolor{weakgray}{80.49} & \textcolor{weakgray}{4.63} & \textcolor{weakgray}{1150} \\
        \hline
        \multirow{5}{*}{\rotatebox[origin=c]{90}{K = 0.4}}
        & Random Drop & 74.57 & 2.52 & - \\
        & Top-K~\cite{liang2022not} & 75.45 & 2.38 & 2143 \\
        & Zero-TP~\cite{wang2024zero} & 75.91 & 2.54 & 1605 \\
        & DynamicViT~\cite{rao2021dynamicvit} & 76.40 & 2.58 & 1934 \\
        & EViT~\cite{liang2022not} & 74.75 & 2.40 & 2112 \\
        & \cellcolor{lavender}TNT (ours) & \cellcolor{lavender}77.32 & \cellcolor{lavender}2.52 & \cellcolor{lavender}1994 \\
        \hline

        \multirow{5}{*}{\rotatebox[origin=c]{90}{K = 0.35}}
        & Random Drop & 72.79 & 2.35 & - \\
        & Top-K~\cite{liang2022not} & 73.87 & 2.20 & 2342 \\
        & Zero-TP~\cite{wang2024zero} & 74.58 & 2.37 & 1679 \\
        & DynamicViT~\cite{rao2021dynamicvit} & 75.22 & 2.41 & 2091 \\
        & EViT~\cite{liang2022not} & 73.73 & 2.21 & 2291 \\
        & \cellcolor{lavender}TNT (ours) & \cellcolor{lavender}76.19 & \cellcolor{lavender}2.35 & \cellcolor{lavender}2167 \\
        \hline

        \multirow{5}{*}{\rotatebox[origin=c]{90}{K = 0.3}}
        & Random Drop & 70.42 & 2.18 & - \\
        & Top-K~\cite{liang2022not} & 71.60 & 2.02 & 2465 \\
        & Zero-TP~\cite{wang2024zero} & 72.70 & 2.20 & 1711 \\
        & DynamicViT~\cite{rao2021dynamicvit} & 73.41 & 2.24 & 2189 \\
        & EViT~\cite{liang2022not} & 71.97 & 1.86 & 2430\\
        & \cellcolor{lavender}TNT (ours) & \cellcolor{lavender}74.65 & \cellcolor{lavender}2.19 & \cellcolor{lavender}2269 \\
        \hline

        \multirow{5}{*}{\rotatebox[origin=c]{90}{K = 0.25}}
        & Random Drop & 67.77 & 2.03 & - \\
        & Top-K~\cite{liang2022not} & 68.57 & 1.86 & 2713 \\
        & Zero-TP~\cite{wang2024zero} & 69.99 & 2.05 & 1757 \\
        & DynamicViT~\cite{rao2021dynamicvit} & 71.03 & 2.08 & 2376 \\
        & EViT~\cite{liang2022not} & 69.66 & 1.70 & 2701 \\
        & \cellcolor{lavender}TNT (ours) & \cellcolor{lavender}72.66 & \cellcolor{lavender}2.03 & \cellcolor{lavender}2444 \\
        \hline

        \multirow{5}{*}{\rotatebox[origin=c]{90}{K = 0.2}}
        & Random Drop & 62.83 & 1.87 & - \\
        & Top-K~\cite{liang2022not} & 64.12 & 1.68 & 2842 \\
        & Zero-TP~\cite{wang2024zero} & 65.90 & 1.88 & 1797 \\
        & DynamicViT~\cite{rao2021dynamicvit} & 66.67 & 1.91 & 2496 \\
        & EViT~\cite{liang2022not} & 66.66 & 1.70 & 2815 \\
        & \cellcolor{lavender}TNT (ours) & \cellcolor{lavender}69.08 & \cellcolor{lavender}1.87 & \cellcolor{lavender}2552 \\
        \hline
    \end{tabular}

    \caption{
    \textbf{Single-layer pruning} for DeiT-S-Distil. in low-token regime.
    }
    \label{tab:deit_s_single_layer_low}
\end{table}

\begin{table*}[h!]
    \centering
    \setlength{\tabcolsep}{3pt} 
    \renewcommand{\arraystretch}{1.0} 
    \begin{tabular}{c|lcccc}
        \hline
        \rowcolor{white}
        & \textbf{Method} & \textbf{Acc.} & \textbf{GFLOPs} & \textbf{TP (imgs/s)} & \textbf{Param.} \\
        \hline
        & \textcolor{weakgray}{Deit-S-Distil.} & \textcolor{weakgray}{80.49} & \textcolor{weakgray}{4.63} & \textcolor{weakgray}{1150} & \textcolor{weakgray}{-}\\
        \hline
        \multirow{5}{*}{\rotatebox[origin=c]{90}{$\tiny\text{GFLOPs } \text{\tiny $ \approx 3.45$}$}}
        & Top-K~\cite{liang2022not} & 79.81 & 3.44 & 1496 & Rate=[.9, .9, .8] \\ 
        & Zero-TP~\cite{wang2024zero} & 79.66 & 3.42 & 983 & Rate=[1., .9, .9, .9, 1.] \\
        & DynamicViT~\cite{rao2021dynamicvit} & 79.45 & 3.47 & 1448 & $\rho$=.8 \\
        & \cellcolor{lavender}ToMe~\cite{bolya2022token} & \cellcolor{lavender}80.12 & \cellcolor{lavender}3.45 & \cellcolor{lavender}1281 & \cellcolor{lavender}$r$=8 \\
        & EViT~\cite{liang2022not} & 79.16 & 3.48 & 1463 & $\rho$=.2 \\
        & TNT (ours) & 79.89 & 3.41 & 1444 & Rate=[1., .95, .95] \\
        \hline

        \multirow{5}{*}{\rotatebox[origin=c]{90}{$\tiny\text{GFLOPs } \text{\tiny $ \approx 3.0$}$}}
        & Top-K~\cite{liang2022not} & 79.11 & 3.06 & 1706 & Rate=[.85, .8, .8] \\
        & Zero-TP~\cite{wang2024zero} & 78.75 & 2.97 & 1171 & Rate=[1., .8, .8, .9, 1.] \\
        & DynamicViT~\cite{rao2021dynamicvit} & 79.18 & 2.99 & 1679 & $\rho$=.7 \\
        & \cellcolor{lavender}ToMe~\cite{bolya2022token} & \cellcolor{lavender}79.86 & \cellcolor{lavender}3.02 & \cellcolor{lavender}1448 & \cellcolor{lavender}$r$=11 \\
        & EViT~\cite{liang2022not} & 79.04 & 3.04 & 1674 & $\rho$=.2 \\
        & TNT (ours) & 79.38 & 2.97 & 1652 & Rate=[.9, .9, .9] \\
        \hline

        \multirow{5}{*}{\rotatebox[origin=c]{90}{$\tiny\text{GFLOPs } \text{\tiny $ \approx 2.6$}$}}
        & Top-K~\cite{liang2022not} & 77.65 & 2.67 & 1909 & Rate=[.85, .7, .7] \\
        & Zero-TP~\cite{wang2024zero} & 76.92 & 2.54 & 1352 & Rate=[1., .7, .7, .8, 1.] \\
        & DynamicViT~\cite{rao2021dynamicvit} & 78.38 & 2.57 & 1929 & $\rho$=.6 \\
        & \cellcolor{lavender}ToMe~\cite{bolya2022token} & \cellcolor{lavender}79.12 & \cellcolor{lavender}2.54 & \cellcolor{lavender}1754 & \cellcolor{lavender}$r$=15 \\
        & EViT~\cite{liang2022not} & 77.60 & 2.64 & 1906 & $\rho$=.2 \\
        & TNT (ours) & 78.57 & 2.60 & 1882 & Rate=[.85, .85, .8] \\
        \hline

        \multirow{4}{*}{\rotatebox[origin=c]{90}{$\tiny\text{GFLOPs } \text{\tiny $ \approx 2.25$}$}}
        & Top-K~\cite{liang2022not} & 75.33 & 2.29 & 2190 & Rate=[.7, .7, .65] \\
        & Zero-TP~\cite{wang2024zero} & 74.35 & 2.25 & 1389 & Rate=[1., .6, .7, .7, 1.] \\
        & DynamicViT~\cite{rao2021dynamicvit} & 76.39 & 2.23 & 2192 & $\rho$=.5 \\
        & \cellcolor{lavender}EViT~\cite{liang2022not} & \cellcolor{lavender}77.97 & \cellcolor{lavender}2.31 & \cellcolor{lavender}2146 & \cellcolor{lavender}$\rho$=.2 \\
        & TNT (ours) & 77.25 & 2.28 & 2143 & Rate=[.75, .85, .7] \\
        \hline

        \multirow{4}{*}{\rotatebox[origin=c]{90}{$\tiny\text{GFLOPs } \text{\tiny $ \approx 2.1$}$}}
        & Top-K~\cite{liang2022not} & 73.73 & 2.13 & 2343 & Rate=[.65, .65, .65] \\
        & Zero-TP~\cite{wang2024zero} & 72.55 & 2.16 & 1415 & Rate=[1., .6, .6, .7, 1.] \\
        & DynamicViT~\cite{rao2021dynamicvit} & 74.41 & 2.08 & 2321 & $\rho$=.45 \\
        & \cellcolor{lavender}EViT~\cite{liang2022not} & \cellcolor{lavender}77.06 & \cellcolor{lavender}2.05 & \cellcolor{lavender}2373 & \cellcolor{lavender}$\rho$=.2 \\
        & TNT (ours) & 75.93 & 2.09 & 2302 & Rate=[.65, .85, .7] \\
        \hline
    \end{tabular}

    \caption{
    \textbf{Multi-layer pruning} for DeiT-S-Distil. in high-token regime.
    }
    \label{tab:deit_s_multi_layer_high}
\end{table*}

\begin{table*}[h!]
    \centering
    \setlength{\tabcolsep}{3pt} 
    \renewcommand{\arraystretch}{1.0} 
    \begin{tabular}{c|lcccc}
        \hline
        \rowcolor{white}
        & \textbf{Method} & \textbf{Acc.} & \textbf{GFLOPs} & \textbf{TP (imgs/s)} & \textbf{Param.} \\
        \hline
        & \textcolor{weakgray}{Deit-S-Distil.} & \textcolor{weakgray}{80.49} & \textcolor{weakgray}{4.63} & \textcolor{weakgray}{1150} & \textcolor{weakgray}{-}\\
        \hline
        \multirow{5}{*}{\rotatebox[origin=c]{90}{$\tiny\text{GFLOPs } \text{\tiny $ \approx 1.95$}$}}
        & Top-K & 70.78 & 1.93 & 2525 & Rate=[.6, .6, .6] \\ 
        & Zero-TP~\cite{wang2024zero} & 69.48 & 2.03 & 1357 & Rate=[1., .5, .5, .6, 1.] \\
        & DynamicViT~\cite{rao2021dynamicvit} & 71.69 & 1.95 & 2490 & $\rho$=.4 \\
        & ToMe~\cite{bolya2022token} & 74.76 & 1.90 & 2176 & $r$=20 \\
        & \cellcolor{lavender}EViT~\cite{liang2022not} & \cellcolor{lavender}76.10 & \cellcolor{lavender}1.93 & \cellcolor{lavender}2494 & \cellcolor{lavender}$\rho$=.2 \\
        & TNT (ours) & 74.54 & 1.95 & 2484 & Rate=[.6, .8, .7] \\
        \hline

        \multirow{5}{*}{\rotatebox[origin=c]{90}{$\tiny\text{GFLOPs } \text{\tiny $ \approx 1.8$}$}}
        & Top-K & 67.34 & 1.80 & 2701 & Rate=[.6, .5, .6] \\ 
        & Zero-TP~\cite{wang2024zero} & 59.29 & 1.85 & 1331 & Rate=[1., .4, .4, .4, 1.] \\
        & DynamicViT~\cite{rao2021dynamicvit} & 67.03 & 1.83 & 2635 & $\rho$=.35 \\
        & ToMe~\cite{bolya2022token} & 71.50 & 1.74 & 2351 & $r$=21 \\
        & \cellcolor{lavender}EViT~\cite{liang2022not} & \cellcolor{lavender}74.27 & \cellcolor{lavender}1.82 & \cellcolor{lavender}2603 & \cellcolor{lavender}$\rho$=.2 \\
        & TNT (ours) & 72.26 & 1.79 & 2639 & Rate=[.5, .8, .7] \\
        \hline

        \multirow{5}{*}{\rotatebox[origin=c]{90}{$\tiny\text{GFLOPs } \text{\tiny $ \approx 1.75$}$}}
        & Top-K & 65.55 & 1.73 & 2794 & Rate=[.55, .5, .6] \\ 
        & Zero-TP~\cite{wang2024zero} & 49.70 & 1.79 & 1471 & Rate=[1., .4, .3, .4, 1.] \\
        & DynamicViT~\cite{rao2021dynamicvit} & 58.70 & 1.71 & 2761 & $\rho$=.3 \\
        & ToMe~\cite{bolya2022token} & 69.76 & 1.68 & 2377 & $r$=22 \\ 
        & \cellcolor{lavender}EViT~\cite{liang2022not} & \cellcolor{lavender}71.53 & \cellcolor{lavender}1.73 & \cellcolor{lavender}2765 & \cellcolor{lavender}$\rho$=.2 \\
        & TNT (ours) & 70.31 & 1.70 & 2725 & Rate=[.5, .7, .7] \\
        \hline

        \multirow{5}{*}{\rotatebox[origin=c]{90}{$\tiny\text{GFLOPs } \text{\tiny $ \approx 1.65$}$}}
        & Top-K & 62.41 & 1.66 & 2896 & Rate=[.55, .5, .5] \\ 
        & Zero-TP~\cite{wang2024zero} & 48.44 & 1.68 & 1481 & Rate=[1., .3, .4, .4, 1.] \\
        & DynamicViT~\cite{rao2021dynamicvit} & 47.71 & 1.62 & 2943 & $\rho$=.25 \\
        & ToMe~\cite{bolya2022token} & 66.30 & 1.61 & 2357 & $r$=24 \\ 
        & EViT~\cite{liang2022not} & 66.97 & 1.65 & 2861 & $\rho$=.2 \\
        & \cellcolor{lavender}TNT (ours) & \cellcolor{lavender}68.36 & \cellcolor{lavender}1.64 & \cellcolor{lavender}2850 & \cellcolor{lavender}Rate=[.5, .7, .6] \\
        \hline

        \multirow{5}{*}{\rotatebox[origin=c]{90}{$\tiny\text{GFLOPs } \text{\tiny $ \approx 1.55$}$}}
        & Top-K & 58.20 & 1.57 & 3003 & Rate=[.5, .45, .5] \\ 
        & Zero-TP~\cite{wang2024zero} & 27.75 & 1.63 & 1485 & Rate=[1., .3, .3, .4, 1.] \\
        & DynamicViT~\cite{rao2021dynamicvit} & 24.51 & 1.52 & 3037 & $\rho$=.2 \\
        & ToMe~\cite{bolya2022token} & 62.88 & 1.55 & 2486 & $r$=25 \\ 
        & EViT~\cite{liang2022not} & 57.25 & 1.57 & 2980 & $\rho$=.2 \\
        & \cellcolor{lavender}TNT (ours) & \cellcolor{lavender}63.82 & \cellcolor{lavender}1.53 & \cellcolor{lavender}2973 & \cellcolor{lavender}Rate=[.5, .6, .55] \\
        \hline
    \end{tabular}

    \caption{
    \textbf{Multi-layer pruning} for DeiT-S-Distil. in low-token regime.
    }
    \label{tab:deit_s_multi_layer_low}
\end{table*}

\subsection{Base Model: DeiT-Tiny-Distil.}

\begin{figure*}[h!]
    \centering
    \includegraphics[width=0.8\textwidth]{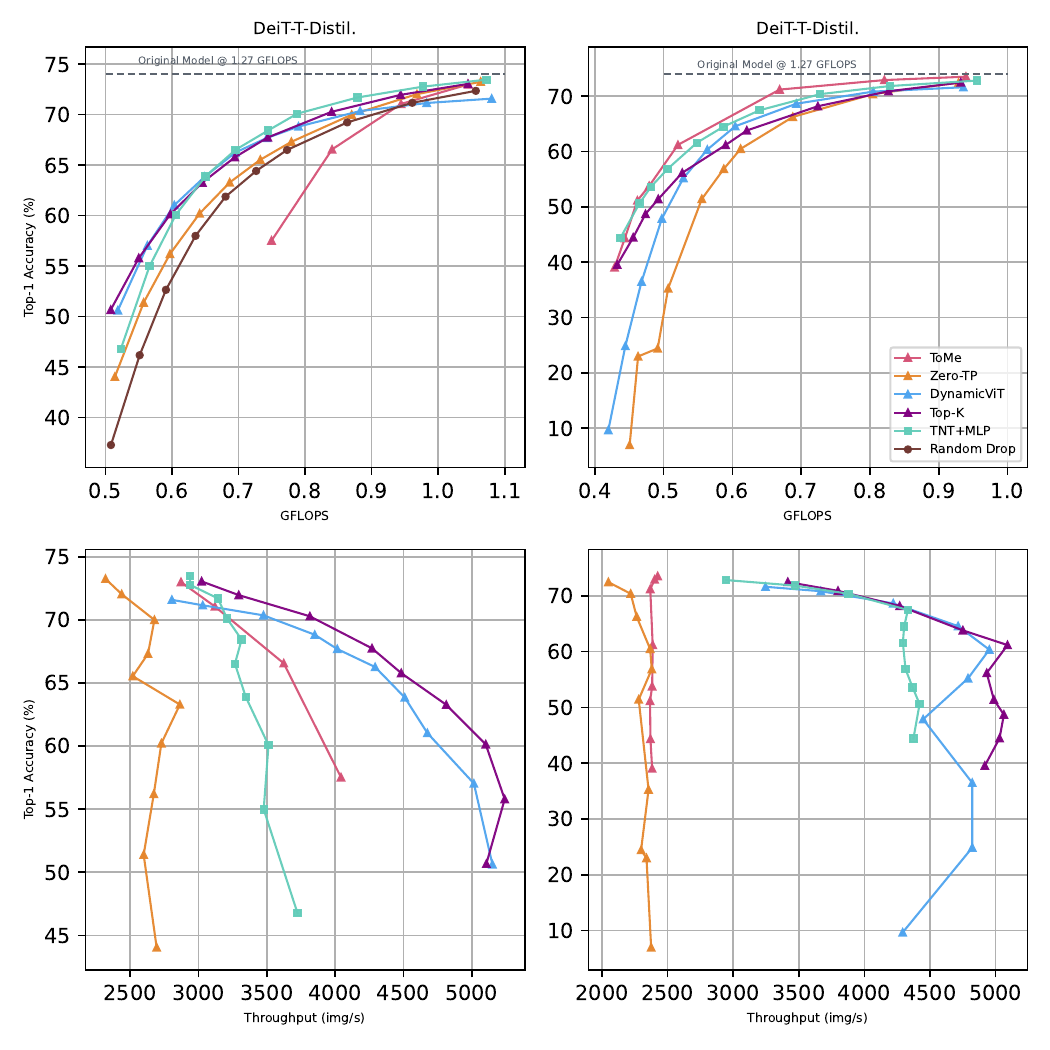}  
    \caption{\textbf{Experimental Results for DeiT-Tiny-Distil.:}
    We plot the Top-1 Accuracy in the ImageNet-1k validation set for each of the pruning methods as a function of computational efficiency, in the \textbf{top row} measured by GFLOPs and in the \textbf{bottom row} measured by throughput, for \textbf{multi-layer pruning}. For weaker encoders and/or smaller transformer blocks such as DeiT-Tiny, we rather use MLP, which means less parallelism. The throughput of models in multi-layer pruning reaches approximately 5000 images per second, potentially limited by hardware bottlenecks.
    }
    \label{fig:deit_t_plot}
\end{figure*}

Figure~\ref{fig:deit_t_plot} shows the plots for DeiT-Tiny-Distil.. For single-layer pruning, high-token regime results are listed in Table~\ref{tab:deit_t_single_layer_high}; low-token regime results are listed in Table~\ref{tab:deit_t_single_layer_low}. For multi-layer pruning, high-token regime results are listed in Table~\ref{tab:deit_t_multi_layer_high}; low-token regime results are listed in Table~\ref{tab:deit_t_multi_layer_low}.

\begin{table}[h!]
    \centering
    \setlength{\tabcolsep}{3pt} 
    \renewcommand{\arraystretch}{1.0} 
    \begin{tabular}{c|lccc}
        \hline
        \rowcolor{white}
        & \textbf{Method} & \textbf{Acc.} & \textbf{GFLOPs} & \textbf{TP (imgs/s)} \\
        \hline
        & \textcolor{weakgray}{Deit-T-Distil.} & \textcolor{weakgray}{74.05} & \textcolor{weakgray}{1.27} & \textcolor{weakgray}{2607} \\
        \hline
        \multirow{6}{*}{\rotatebox[origin=c]{90}{K = 0.8}}
        & Random Drop & 72.36 & 1.06 & - \\
        & Top-K~\cite{liang2022not} & 73.05 & 1.04 & 3023 \\
        & Zero-TP~\cite{wang2024zero} & 73.25 & 1.06 & 2319 \\
        & DynamicViT~\cite{rao2021dynamicvit} & 71.59 & 1.08 & 2805 \\
        & ToMe~\cite{bolya2022token} & 72.99 & 1.05 & 2872 \\
        & \cellcolor{lavender}TNT+MLP (ours) & \cellcolor{lavender}73.46 & \cellcolor{lavender}1.07 & \cellcolor{lavender}2938 \\
        \hline

        \multirow{6}{*}{\rotatebox[origin=c]{90}{K = 0.7}}
        & Random Drop & 71.17 & 0.96 & - \\
        & Top-K~\cite{liang2022not} & 71.96 & 0.94 & 3295 \\
        & Zero-TP~\cite{wang2024zero} & 72.03 & 0.97 & 2438 \\
        & DynamicViT~\cite{rao2021dynamicvit} & 71.15 & 0.98 & 3031 \\
        & ToMe~\cite{bolya2022token} & 71.06 & 0.94 & 3120 \\
        & \cellcolor{lavender}TNT+MLP (ours) & \cellcolor{lavender}72.77 & \cellcolor{lavender}0.98 & \cellcolor{lavender}2939 \\
        \hline

        \multirow{6}{*}{\rotatebox[origin=c]{90}{K = 0.6}}
        & Random Drop & 69.23 & 0.86 & - \\
        & Top-K~\cite{liang2022not} & 70.28 & 0.84 & 3815 \\
        & Zero-TP~\cite{wang2024zero} & 69.99 & 0.87 & 2677 \\
        & DynamicViT~\cite{rao2021dynamicvit} & 70.36 & 0.88 & 3476 \\
        & ToMe~\cite{bolya2022token} & 66.56 & 0.84 & 3624 \\
        & \cellcolor{lavender}TNT+MLP (ours) & \cellcolor{lavender}71.70 & \cellcolor{lavender}0.88 & \cellcolor{lavender}3142 \\
        \hline

        \multirow{6}{*}{\rotatebox[origin=c]{90}{K = 0.5}}
        & Random Drop & 66.50 & 0.77 & - \\
        & Top-K~\cite{liang2022not} & 67.74 & 0.74 & 4270 \\
        & Zero-TP~\cite{wang2024zero} & 67.31 & 0.78 & 2630 \\
        & DynamicViT~\cite{rao2021dynamicvit} & 68.82 & 0.79 & 3851 \\
        & ToMe~\cite{bolya2022token} & 57.52 & 0.75 & 4043 \\
        & \cellcolor{lavender}TNT+MLP (ours) & \cellcolor{lavender}70.09 & \cellcolor{lavender}0.79 & \cellcolor{lavender}3210 \\
        \hline

        \multirow{5}{*}{\rotatebox[origin=c]{90}{K = 0.45}}
        & Random Drop & 64.43 & 0.73 & - \\
        & Top-K~\cite{liang2022not} & 65.78 & 0.69 & 4484 \\
        & Zero-TP~\cite{wang2024zero} & 65.54 & 0.73 & 2519 \\
        & DynamicViT~\cite{rao2021dynamicvit} & 67.69 & 0.74 & 4016 \\
        & \cellcolor{lavender}TNT+MLP (ours) & \cellcolor{lavender}68.43 & \cellcolor{lavender}0.74 & \cellcolor{lavender}3315 \\
        \hline
    \end{tabular}

    \caption{
    \textbf{Single-layer pruning} for DeiT-T-Distil. in low-token regime.
    }
    \label{tab:deit_t_single_layer_high}
\end{table}

\begin{table}[h!]
    \centering
    \setlength{\tabcolsep}{3pt} 
    \renewcommand{\arraystretch}{1.0} 
    \begin{tabular}{c|lccc}
        \hline
        \rowcolor{white}
        & \textbf{Method} & \textbf{Acc.} & \textbf{GFLOPs} & \textbf{TP (imgs/s)} \\
        \hline
        & \textcolor{weakgray}{Deit-T-Distil.} & \textcolor{weakgray}{74.05} & \textcolor{weakgray}{1.27} & \textcolor{weakgray}{2607} \\
        \hline
        \multirow{5}{*}{\rotatebox[origin=c]{90}{K = 0.4}}
        & Random Drop & 61.88 & 0.68 & - \\
        & Top-K~\cite{liang2022not} & 63.27 & 0.65 & 4814 \\
        & Zero-TP~\cite{wang2024zero} & 63.29 & 0.69 & 2864 \\
        & DynamicViT~\cite{rao2021dynamicvit} & 66.26 & 0.70 & 4293 \\
        & \cellcolor{lavender}TNT+MLP (ours) & \cellcolor{lavender}66.52 & \cellcolor{lavender}0.70 & \cellcolor{lavender}3267 \\
        \hline

        \multirow{5}{*}{\rotatebox[origin=c]{90}{K = 0.35}}
        & Random Drop & 57.99 & 0.64 & - \\
        & Top-K~\cite{liang2022not} & 60.13 & 0.60 & 5103 \\
        & Zero-TP~\cite{wang2024zero} & 60.21 & 0.64 & 2728 \\
        & DynamicViT~\cite{rao2021dynamicvit} & 63.85 & 0.65 & 4510 \\
        & \cellcolor{lavender}TNT+MLP (ours) & \cellcolor{lavender}63.89 & \cellcolor{lavender}0.65 & \cellcolor{lavender}3346 \\
        \hline

        \multirow{5}{*}{\rotatebox[origin=c]{90}{K = 0.3}}
        & Random Drop & 52.65 & 0.59 & - \\
        & Top-K~\cite{liang2022not} & 55.79 & 0.55 & 5242 \\
        & Zero-TP~\cite{wang2024zero} & 56.20 & 0.60 & 2673 \\
        & \cellcolor{lavender}DynamicViT~\cite{rao2021dynamicvit} & \cellcolor{lavender}61.03 & \cellcolor{lavender}0.60 & \cellcolor{lavender}4676 \\
        & TNT+MLP (ours) & 60.08 & 0.61 & 3512 \\
        \hline

        \multirow{5}{*}{\rotatebox[origin=c]{90}{K = 0.25}}
        & Random Drop & 46.18 & 0.55 & - \\
        & Top-K~\cite{liang2022not} & 50.67 & 0.51 & 5108 \\
        & Zero-TP~\cite{wang2024zero} & 51.39 & 0.56 & 2598 \\
        & \cellcolor{lavender}DynamicViT~\cite{rao2021dynamicvit} & \cellcolor{lavender}57.03 & \cellcolor{lavender}0.56 & \cellcolor{lavender}5016 \\
        & TNT+MLP (ours) & 54.98 & 0.57 & 3477 \\
        \hline

        \multirow{5}{*}{\rotatebox[origin=c]{90}{K = 0.2}}
        & Random Drop & 37.29 & 0.51 & - \\
        & Top-K~\cite{liang2022not} & 43.40 & 0.46 & 5153 \\
        & Zero-TP~\cite{wang2024zero} & 44.04 & 0.51 & 2694 \\
        & \cellcolor{lavender}DynamicViT~\cite{rao2021dynamicvit} & \cellcolor{lavender}50.62 & \cellcolor{lavender}0.52 & \cellcolor{lavender}5152 \\
        & TNT+MLP (ours) & 46.77 & 0.52 & 3725 \\
        \hline
    \end{tabular}

    \caption{
    \textbf{Single-layer pruning} for DeiT-T-Distil. in low-token regime.
    }
    \label{tab:deit_t_single_layer_low}
\end{table}

\begin{table*}[h!]
    \centering
    \setlength{\tabcolsep}{3pt} 
    \renewcommand{\arraystretch}{1.0} 
    \begin{tabular}{c|lcccc}
        \hline
        \rowcolor{white}
        & \textbf{Method} & \textbf{Acc.} & \textbf{GFLOPs} & \textbf{TP (imgs/s)} & \textbf{Param.} \\
        \hline
        & \textcolor{weakgray}{Deit-T-Distil.} & \textcolor{weakgray}{74.05} & \textcolor{weakgray}{1.27} & \textcolor{weakgray}{2607} & \textcolor{weakgray}{-}\\
        \hline
        \multirow{5}{*}{\rotatebox[origin=c]{90}{$\tiny\text{GFLOPs } \text{\tiny $ \approx 0.93$}$}}
        & Top-K~\cite{liang2022not} & 72.46 & 0.93 & 3416 & Rate=[.9, .9, .8] \\ 
        & Zero-TP~\cite{wang2024zero} & 72.45 & 0.93 & 2048 & Rate=[1., .9, .9, .9, 1.] \\
        & DynamicViT~\cite{rao2021dynamicvit} & 71.61 & 0.94 & 3246 & $\rho$=.8 \\
        & \cellcolor{lavender}ToMe~\cite{bolya2022token} & \cellcolor{lavender}73.52 & \cellcolor{lavender}0.94 & \cellcolor{lavender}2424 & \cellcolor{lavender}$r$=8 \\ 
        & TNT+MLP (ours) & 72.82 & 0.96 & 2947 & Rate=[1., .95, .95] \\
        \hline

        \multirow{5}{*}{\rotatebox[origin=c]{90}{$\tiny\text{GFLOPs } \text{\tiny $ \approx 0.82$}$}}
        & Top-K~\cite{liang2022not} & 70.88 & 0.83 & 3799 & Rate=[.85, .8, .8] \\ 
        & Zero-TP~\cite{wang2024zero} & 70.36 & 0.80 & 2219 & Rate=[1., .8, .8, .9, 1.] \\
        & DynamicViT~\cite{rao2021dynamicvit} & 70.82 & 0.80 & 3668 & $\rho$=.7 \\
        & \cellcolor{lavender}ToMe~\cite{bolya2022token} & \cellcolor{lavender}72.91 & \cellcolor{lavender}0.82 & \cellcolor{lavender}2401 & \cellcolor{lavender}$r$=11 \\ 
        & TNT+MLP (ours) & 71.83 & 0.83 & 3470 & Rate=[.9, .9, .9] \\
        \hline

        \multirow{5}{*}{\rotatebox[origin=c]{90}{$\tiny\text{GFLOPs } \text{\tiny $ \approx 0.7$}$}}
        & Top-K~\cite{liang2022not} & 68.18 & 0.72 & 4269 & Rate=[.85, .7, .7] \\ 
        & Zero-TP~\cite{wang2024zero} & 66.24 & 0.69 & 2264 & Rate=[1., .7, .7, .8, 1.] \\
        & DynamicViT~\cite{rao2021dynamicvit} & 68.63 & 0.69 & 4220 & $\rho$=.6 \\
        & \cellcolor{lavender}ToMe~\cite{bolya2022token} & \cellcolor{lavender}71.19 & \cellcolor{lavender}0.67 & \cellcolor{lavender}2368 & \cellcolor{lavender}$r$=15 \\ 
        & TNT+MLP (ours) & 70.36 & 0.73 & 3879 & Rate=[.85, .85, .8] \\
        \hline

        \multirow{4}{*}{\rotatebox[origin=c]{90}{$\tiny\text{GFLOPs } \text{\tiny $ \approx 0.62$}$}}
        & Top-K~\cite{liang2022not} & 63.80 & 0.62 & 4751 & Rate=[.7, .7, .65] \\ 
        & Zero-TP~\cite{wang2024zero} & 60.48 & 0.61 & 2367 & Rate=[1., .6, .7, .7, 1.] \\
        & DynamicViT~\cite{rao2021dynamicvit} & 64.55 & 0.60 & 4715 & $\rho$=.5 \\
        & \cellcolor{lavender}TNT+MLP (ours) & \cellcolor{lavender}67.45 & \cellcolor{lavender}0.64 & \cellcolor{lavender}4332 & \cellcolor{lavender}Rate=[.75, .85, .7] \\
        \hline

        \multirow{4}{*}{\rotatebox[origin=c]{90}{$\tiny\text{GFLOPs } \text{\tiny $ \approx 0.58$}$}}
        & Top-K~\cite{liang2022not} & 61.17 & 0.59 & 5092 & Rate=[.65, .65, .65] \\ 
        & Zero-TP~\cite{wang2024zero} & 56.84 & 0.59 & 2380 & Rate=[1., .6, .6, .7, 1.] \\
        & DynamicViT~\cite{rao2021dynamicvit} & 60.34 & 0.56 & 4953 & $\rho$=.45 \\
        & \cellcolor{lavender}TNT+MLP (ours) & \cellcolor{lavender}64.48 & \cellcolor{lavender}0.59 & \cellcolor{lavender}4304 & \cellcolor{lavender}Rate=[.65, .85, .7] \\
        \hline
    \end{tabular}

    \caption{
    \textbf{Multi-layer pruning} for DeiT-T-Distil. in high-token regime.
    }
    \label{tab:deit_t_multi_layer_high}
\end{table*}

\begin{table*}[h!]
    \centering
    \setlength{\tabcolsep}{3pt} 
    \renewcommand{\arraystretch}{1.0} 
    \begin{tabular}{c|lcccc}
        \hline
        \rowcolor{white}
        & \textbf{Method} & \textbf{Acc.} & \textbf{GFLOPs} & \textbf{TP (imgs/s)} & \textbf{Param.} \\
        \hline
        & \textcolor{weakgray}{Deit-T-Distil.} & \textcolor{weakgray}{74.05} & \textcolor{weakgray}{1.27} & \textcolor{weakgray}{2607} & \textcolor{weakgray}{-}\\
        \hline
        \multirow{5}{*}{\rotatebox[origin=c]{90}{$\tiny\text{GFLOPs } \text{\tiny $ \approx 0.53$}$}}
        & Top-K~\cite{liang2022not} & 56.15 & 0.53 & 4933 & Rate=[.6, .6, .6] \\ 
        & Zero-TP~\cite{wang2024zero} & 51.41 & 0.56 & 2280 & Rate=[1., .5, .5, .6, 1.] \\ 
        & DynamicViT~\cite{rao2021dynamicvit} & 55.20 & 0.53 & 4790 & $\rho$=.4 \\ 
        & ToMe~\cite{bolya2022token} & 61.20 & 0.52 & 2386 & $r$=20 \\ 
        & \cellcolor{lavender}TNT+MLP (ours) & \cellcolor{lavender}61.59 & \cellcolor{lavender}0.55 & \cellcolor{lavender}4294 & \cellcolor{lavender}Rate=[.6, .8, .7] \\
        \hline

        \multirow{5}{*}{\rotatebox[origin=c]{90}{$\tiny\text{GFLOPs } \text{\tiny $ \approx 0.5$}$}}
        & Top-K~\cite{liang2022not} & 51.39 & 0.49 & 4987 & Rate=[.6, .5, .6] \\ 
        & Zero-TP~\cite{wang2024zero} & 35.26 & 0.51 & 2355 & Rate=[1., .4, .4, .4, 1.] \\ 
        & DynamicViT~\cite{rao2021dynamicvit} & 47.87 & 0.50 & 4448 & $\rho$=.35 \\ 
        & ToMe~\cite{bolya2022token} & 53.77 & 0.48 & 2382 & $r$=21 \\ 
        & \cellcolor{lavender}TNT+MLP (ours) & \cellcolor{lavender}56.85 & \cellcolor{lavender}0.51 & \cellcolor{lavender}4314 & \cellcolor{lavender}Rate=[.5, .8, .7] \\
        \hline

        \multirow{5}{*}{\rotatebox[origin=c]{90}{$\tiny\text{GFLOPs } \text{\tiny $ \approx 0.47$}$}}
        & Top-K~\cite{liang2022not} & 48.70 & 0.47 & 5064 & Rate=[.55, .5, .6] \\ 
        & Zero-TP~\cite{wang2024zero} & 24.42 & 0.49 & 2298 & Rate=[1., .4, .3, .4, 1.] \\ 
        & DynamicViT~\cite{rao2021dynamicvit} & 36.48 & 0.47 & 4823 & $\rho$=.3 \\ 
        & ToMe~\cite{bolya2022token} & 51.15 & 0.46 & 2366 & $r$=22 \\ 
        & \cellcolor{lavender}TNT+MLP (ours) & \cellcolor{lavender}53.58 & \cellcolor{lavender}0.48 & \cellcolor{lavender}4368 & \cellcolor{lavender}Rate=[.5, .7, .7] \\
        \hline

        \multirow{5}{*}{\rotatebox[origin=c]{90}{$\tiny\text{GFLOPs } \text{\tiny $ \approx 0.45$}$}}
        & Top-K~\cite{liang2022not} & 44.47 & 0.46 & 5031 & Rate=[.55, .5, .5] \\ 
        & Zero-TP~\cite{wang2024zero} & 22.97 & 0.46 & 2339 & Rate=[1., .3, .4, .4, 1.] \\ 
        & DynamicViT~\cite{rao2021dynamicvit} & 24.86 & 0.44 & 4823 & $\rho$=.25 \\ 
        & ToMe~\cite{bolya2022token} & 44.34 & 0.44 & 2369 & $r$=24 \\ 
        & \cellcolor{lavender}TNT+MLP (ours) & \cellcolor{lavender}50.60 & \cellcolor{lavender}0.46 & \cellcolor{lavender}4421 & \cellcolor{lavender}Rate=[.5, .7, .6] \\
        \hline

        \multirow{5}{*}{\rotatebox[origin=c]{90}{$\tiny\text{GFLOPs } \text{\tiny $ \approx 0.43$}$}}
        & Top-K~\cite{liang2022not} & 39.54 & 0.43 & 4919 & Rate=[.5, .45, .5] \\ 
        & Zero-TP~\cite{wang2024zero} & 6.97 & 0.45 & 2374 & Rate=[1., .3, .3, .4, 1.] \\ 
        & DynamicViT~\cite{rao2021dynamicvit} & 9.65 & 0.42 & 4292 & $\rho$=.2 \\ 
        & ToMe~\cite{bolya2022token} & 39.05 & 0.43 & 2382 & $r$=25 \\ 
        & \cellcolor{lavender}TNT+MLP (ours) & \cellcolor{lavender}44.41 & \cellcolor{lavender}0.44 & \cellcolor{lavender}4374 & \cellcolor{lavender}Rate=[.5, .6, .55] \\
        \hline
    \end{tabular}

    \caption{
    \textbf{Multi-layer pruning} for DeiT-T-Distil. in low-token regime.
    }
    \label{tab:deit_t_multi_layer_low}
\end{table*}

\subsection{Base Model: ViT/16}
For ViT/16, we use the mean-pooled tokens for the prediction rather than using \texttt{CLS} token. Therefore, Top-K is not applicable. We instead sweep the number of tokens (denoted as \textbf{\#tokens}) from 160 to 50. \#Tokens from 160 to 110 represent the high-token regime while \#tokens from 100 to 50 correspond to low-token regime. For single-layer pruning, high-token regime results are listed in Table~\ref{tab:vit16_single_layer_high}; low-token regime results are listed in Table~\ref{tab:vit16_single_layer_low}. For multi-layer pruning, high-token regime results are listed in Table~\ref{tab:vit16_multi_layer_high}; low-token regime results are listed in Table~\ref{tab:vit16_multi_layer_low}.

\begin{table}[h!]
    \centering
    \setlength{\tabcolsep}{3pt} 
    \renewcommand{\arraystretch}{1.0} 
    \begin{tabular}{c|lccc}
        \hline
        \rowcolor{white}
        \textbf{\#Tokens} & \textbf{Method} & \textbf{Acc.} & \textbf{GFLOPs} & \textbf{TP (imgs/s)} \\
        \hline
        & \textcolor{weakgray}{ViT/16} & \textcolor{weakgray}{78.70} & \textcolor{weakgray}{9.17} & \textcolor{weakgray}{644} \\
        \hline
        \multirow{4}{*}{160}
        & Random Drop & 77.17 & 7.69 & - \\
        & Zero-TP~\cite{wang2024zero} & 77.43 & 7.32 & 665 \\
        & ToMe~\cite{bolya2022token} & 77.71 & 7.66 & 752 \\
        & DynamicViT~\cite{rao2021dynamicvit} & 77.18 & 7.39 & 723 \\
        & \cellcolor{lavender}TNT (ours) & \cellcolor{lavender}77.94 & \cellcolor{lavender}7.70 & \cellcolor{lavender}747 \\
        \hline

        \multirow{4}{*}{150}
        & Random Drop & 76.75 & 7.29 & - \\
        & Zero-TP~\cite{wang2024zero} & 76.80 & 6.92 & 687 \\
        & ToMe~\cite{bolya2022token} & 77.22 & 7.24 & 784 \\
        & DynamicViT~\cite{rao2021dynamicvit} & 76.99 & 7.98 & 749 \\
        & \cellcolor{lavender}TNT (ours) & \cellcolor{lavender}77.64 & \cellcolor{lavender}7.30 & \cellcolor{lavender}779 \\
        \hline

        \multirow{4}{*}{140}
        & Random Drop & 76.00 & 6.89 & - \\
        & Zero-TP~\cite{wang2024zero} & 76.80 & 6.92 & 722 \\
        & ToMe~\cite{bolya2022token} & 76.53 & 6.83 & 838 \\
        & DynamicViT~\cite{rao2021dynamicvit} & 76.12 & 7.54 & 780 \\
        & \cellcolor{lavender}TNT (ours) & \cellcolor{lavender}76.97 & \cellcolor{lavender}6.90 & \cellcolor{lavender}820 \\
        \hline

        \multirow{4}{*}{130}
        & Random Drop & 75.12 & 6.50 & - \\
        & Zero-TP~\cite{wang2024zero} & 75.96 & 6.52 & 749 \\
        & ToMe~\cite{bolya2022token} & 75.70 & 6.42 & 877 \\
        & DynamicViT~\cite{rao2021dynamicvit} & 75.25 & 7.02 & 826 \\
        & \cellcolor{lavender}TNT (ours) & \cellcolor{lavender}76.22 & \cellcolor{lavender}6.50 & \cellcolor{lavender}856 \\
        \hline

        \multirow{4}{*}{120}
        & Random Drop & 73.79 & 6.10 & - \\
        & Zero-TP~\cite{wang2024zero} & 74.81 & 6.13 & 812 \\
        & ToMe~\cite{bolya2022token} & 74.19 & 6.02 & 969 \\
        & DynamicViT~\cite{rao2021dynamicvit} & 74.51 & 6.68 & 882 \\
        & \cellcolor{lavender}TNT (ours) & \cellcolor{lavender}75.24 & \cellcolor{lavender}6.11 & \cellcolor{lavender}945 \\
        \hline

        \multirow{4}{*}{110}
        & Random Drop & 71.95 & 5.71 & - \\
        & Zero-TP~\cite{wang2024zero} & 73.26 & 5.74 & 863 \\
        & ToMe~\cite{bolya2022token} & 71.54 & 5.62 & 1047 \\
        & DynamicViT~\cite{rao2021dynamicvit} & 72.88 & 6.26 & 941 \\
        & \cellcolor{lavender}TNT (ours) & \cellcolor{lavender}73.73 & \cellcolor{lavender}5.72 & \cellcolor{lavender}1016 \\
        \hline
    \end{tabular}

    \caption{
    \textbf{Single-layer pruning} for ViT/16. in high-token regime.
    }
    \label{tab:vit16_single_layer_high}
\end{table}

\begin{table}[h!]
    \centering
    \setlength{\tabcolsep}{3pt} 
    \renewcommand{\arraystretch}{1.0} 
    \begin{tabular}{c|lccc}
        \hline
        \rowcolor{white}
        \textbf{\#Tokens} & \textbf{Method} & \textbf{Acc.} & \textbf{GFLOPs} & \textbf{TP (imgs/s)} \\
        \hline
        & \textcolor{weakgray}{ViT/16} & \textcolor{weakgray}{78.70} & \textcolor{weakgray}{9.17} & \textcolor{weakgray}{644} \\
        \hline
        \multirow{3}{*}{100}
        & Random Drop & 69.21 & 5.33 & - \\
        & Zero-TP~\cite{wang2024zero} & 70.92 & 5.35 & 901 \\
        & ToMe~\cite{bolya2022token} & 66.46 & 5.22 & 1103 \\
        & DynamicViT~\cite{rao2021dynamicvit} & 70.77 & 5.81 & 997 \\
        & \cellcolor{lavender}TNT (ours) & \cellcolor{lavender}71.69 & \cellcolor{lavender}5.33 & \cellcolor{lavender}1069 \\
        \hline

        \multirow{3}{*}{90}
        & Random Drop & 64.71 & 4.94 & - \\
        & Zero-TP~\cite{wang2024zero} & 67.40 & 4.97 & 960 \\
        & DynamicViT~\cite{rao2021dynamicvit} & 68.03 & 5.43 & 1069 \\
        & \cellcolor{lavender}TNT (ours) & \cellcolor{lavender}68.73 & \cellcolor{lavender}4.95 & \cellcolor{lavender}1153 \\
        \hline

        \multirow{3}{*}{80}
        & Random Drop & 57.64 & 4.56 & - \\
        & Zero-TP~\cite{wang2024zero} & 61.76 & 4.59 & 1002 \\
        & DynamicViT~\cite{rao2021dynamicvit} & 62.27 & 5.04 & 1156 \\
        & \cellcolor{lavender}TNT (ours) & \cellcolor{lavender}63.87 & \cellcolor{lavender}4.56 & \cellcolor{lavender}1247 \\
        \hline

        \multirow{3}{*}{70}
        & Random Drop & 46.28 & 4.18 & - \\
        & Zero-TP~\cite{wang2024zero} & 53.15 & 4.21 & 1021 \\
        & DynamicViT~\cite{rao2021dynamicvit} & 54.72 & 4.70 & 1264 \\
        & \cellcolor{lavender}TNT (ours) & \cellcolor{lavender}55.19 & \cellcolor{lavender}4.19 & \cellcolor{lavender}1344 \\
        \hline

        \multirow{3}{*}{60}
        & Random Drop & 31.16 & 3.81 & - \\
        & Zero-TP~\cite{wang2024zero} & 40.84 & 3.83 & 1157 \\
        & DynamicViT~\cite{rao2021dynamicvit} & 39.68 & 4.06 & 1380 \\
        & \cellcolor{lavender}TNT (ours) & \cellcolor{lavender}41.75 & \cellcolor{lavender}3.81 & \cellcolor{lavender}1457 \\
        \hline

        \multirow{3}{*}{50}
        & Random Drop & 16.57 & 3.44 & - \\
        & \cellcolor{lavender}Zero-TP~\cite{wang2024zero} & \cellcolor{lavender}27.64 & \cellcolor{lavender}3.46 & \cellcolor{lavender}1228 \\
        & DynamicViT~\cite{rao2021dynamicvit} & 25.61 & 3.62 & 1492 \\
        & TNT (ours) & 27.40 & 3.44 & 1531 \\
        \hline
    \end{tabular}

    \caption{
    \textbf{Single-layer pruning} for ViT/16. in low-token regime.
    }
    \label{tab:vit16_single_layer_low}
\end{table}

\begin{table*}[h!]
    \centering
    \setlength{\tabcolsep}{3pt} 
    \renewcommand{\arraystretch}{1.7} 
    \begin{tabular}{c|lcccc}
        \hline
        \rowcolor{white}
        & \textbf{Method} & \textbf{Acc.} & \textbf{GFLOPs} & \textbf{TP (imgs/s)} & \textbf{Param.} \\
        \hline
        & \textcolor{weakgray}{ViT/16} & \textcolor{weakgray}{78.70} & \textcolor{weakgray}{9.17} & \textcolor{weakgray}{644} & \textcolor{weakgray}{-}\\
        \hline
        \multirow{3}{*}{\rotatebox[origin=c]{90}{$\tiny\text{GFLOPs } \text{\tiny $ \approx 6.8$}$}}
        & Zero-TP~\cite{wang2024zero} & 77.07 & 6.75 & 688 & Rate=[1., .9, .9, .9, 1.] \\
        & ToMe~\cite{bolya2022token} & 77.02 & 6.97 & 734 & $r$=8 \\
        & DynamicViT~\cite{rao2021dynamicvit} & 78.56 & 7.23 & 831 & $\rho$=.8 \\
        & \cellcolor{lavender}TNT (ours) & \cellcolor{lavender}77.32 & \cellcolor{lavender}6.73 & \cellcolor{lavender}842 & \cellcolor{lavender}Rate=[1., .95, .95] \\
        \hline

        \multirow{3}{*}{\rotatebox[origin=c]{90}{$\tiny\text{GFLOPs } \text{\tiny $ \approx 6.00$}$}}
        & Zero-TP~\cite{wang2024zero} & 75.16 & 5.84 & 778 & Rate=[1., .8, .8, .9, 1.] \\
        & \cellcolor{lavender}ToMe~\cite{bolya2022token} & \cellcolor{lavender}77.02 & \cellcolor{lavender}6.14 & \cellcolor{lavender}823 & \cellcolor{lavender}$r$=11 \\
        & DynamicViT~\cite{rao2021dynamicvit} & 76.68 & 6.57 & 928 & $\rho$=.7 \\
        & TNT (ours) & 75.47 & 5.84 & 967 & Rate=[.9, .9, .9] \\
        \hline

        \multirow{3}{*}{\rotatebox[origin=c]{90}{$\tiny\text{GFLOPs } \text{\tiny $ \approx 5.00$}$}}
        & Zero-TP~\cite{wang2024zero} & 69.91 & 4.98 & 885 & Rate=[1., .7, .7, .8, 1.] \\
        & ToMe~\cite{bolya2022token} & 70.95 & 5.06 & 989 & $r$=15 \\
        & DynamicViT~\cite{rao2021dynamicvit} & 75.02 & 6.00 & 1011 & $\rho$=.6 \\
        & \cellcolor{lavender}TNT (ours) & \cellcolor{lavender}72.23 & \cellcolor{lavender}5.10 & \cellcolor{lavender}1103 & \cellcolor{lavender}Rate=[.85, .85, .8] \\
        \hline

        \multirow{2}{*}{\rotatebox[origin=c]{90}{$\tiny\text{GFLOPs } \text{\tiny $ \approx 4.43$}$}}
        & Zero-TP~\cite{wang2024zero} & 60.39 & 4.41 & 966 & Rate=[1., .6, .7, .7, 1.] \\
        & DynamicViT~\cite{rao2021dynamicvit} & 71.87 & 5.66 & 1105 & $\rho$=.5 \\
        & \cellcolor{lavender}TNT (ours) & \cellcolor{lavender}65.81 & \cellcolor{lavender}4.46 & \cellcolor{lavender}1249 & \cellcolor{lavender}Rate=[.75, .85, .7] \\
        \hline

        \multirow{2}{*}{\rotatebox[origin=c]{90}{$\tiny\text{GFLOPs } \text{\tiny $ \approx 4.10$}$}}
        & Zero-TP~\cite{wang2024zero} & 51.96 & 4.25 & 994 & Rate=[1., .6, .6, .7, 1.] \\
        & DynamicViT~\cite{rao2021dynamicvit} & 67.63 & 5.17 & 1202 & $\rho$=.45 \\
        & \cellcolor{lavender}TNT (ours) & \cellcolor{lavender}58.96 & \cellcolor{lavender}4.08 & \cellcolor{lavender}1351 & \cellcolor{lavender}Rate=[.65, .85, .7] \\
        \hline
    \end{tabular}

    \caption{
    \textbf{Multi-layer pruning} for ViT/16. in high-token regime.
    }
    \label{tab:vit16_multi_layer_high}
\end{table*}

\begin{table*}[h!]
    \centering
    \setlength{\tabcolsep}{3pt} 
    \renewcommand{\arraystretch}{1.5} 
    \begin{tabular}{c|lcccc}
        \hline
        \rowcolor{white}
        & \textbf{Method} & \textbf{Acc.} & \textbf{GFLOPs} & \textbf{TP (imgs/s)} & \textbf{Param.} \\
        \hline
        & \textcolor{weakgray}{ViT/16} & \textcolor{weakgray}{78.70} & \textcolor{weakgray}{9.17} & \textcolor{weakgray}{644} & \textcolor{weakgray}{-}\\
        \hline
        \multirow{3}{*}{\rotatebox[origin=c]{90}{$\tiny\text{GFLOPs } \text{\tiny $ \approx 3.90$}$}}
        & Zero-TP~\cite{wang2024zero} & 40.38 & 3.98 & 1035 & Rate=[1., .6, .6, .7, 1.] \\
        & ToMe~\cite{bolya2022token} & 41.78 & 3.96 & 1243 & $r$=20 \\
        & DynamicViT~\cite{rao2021dynamicvit} & 60.74 & 4.78 & 1313 & $\rho$=.4 \\
        & \cellcolor{lavender}TNT (ours) & \cellcolor{lavender}51.40 & \cellcolor{lavender}3.80 & \cellcolor{lavender}1452 & \cellcolor{lavender}Rate=[.65, .85, .7] \\
        \hline

        \multirow{3}{*}{\rotatebox[origin=c]{90}{$\tiny\text{GFLOPs } \text{\tiny $ \approx 3.60$}$}}
        & Zero-TP~\cite{wang2024zero} & 13.86 & 3.60 & 1092 & Rate=[1., .5, .5, .6, 1.] \\
        & ToMe~\cite{bolya2022token} & 22.58 & 3.64 & 1340 & $r$=21 \\
        & DynamicViT~\cite{rao2021dynamicvit} & 50.15 & 4.36 & 1417 & $\rho$=.35 \\
        & \cellcolor{lavender}TNT (ours) & \cellcolor{lavender}39.81 & \cellcolor{lavender}3.50 & \cellcolor{lavender}1546 & \cellcolor{lavender}Rate=[.5, .8, .7] \\
        \hline

        \multirow{3}{*}{\rotatebox[origin=c]{90}{$\tiny\text{GFLOPs } \text{\tiny $ \approx 3.40$}$}}
        & Zero-TP~\cite{wang2024zero} & 5.15 & 3.48 & 1102 & Rate=[1., .4, .3, .4, 1.] \\
        & ToMe~\cite{bolya2022token} & 17.74 & 3.51 & 1382 & $r$=21 \\
        & DynamicViT~\cite{rao2021dynamicvit} & 37.04 & 4.03 & 1486 & $\rho$=.3 \\
        & \cellcolor{lavender}TNT (ours) & \cellcolor{lavender}30.86 & \cellcolor{lavender}3.31 & \cellcolor{lavender}1626 & \cellcolor{lavender}Rate=[.5, .7, .7] \\
        \hline

        \multirow{3}{*}{\rotatebox[origin=c]{90}{$\tiny\text{GFLOPs } \text{\tiny $ \approx 3.20$}$}}
        & Zero-TP~\cite{wang2024zero} & 4.12 & 3.27 & 1163 & Rate=[1., .3, .4, .4, 1.] \\
        & ToMe~\cite{bolya2022token} & 9.19 & 3.37 & 1446 & $r$=24 \\
        & DynamicViT~\cite{rao2021dynamicvit} & 24.91 & 3.96 & 1609 & $\rho$=.25 \\
        & \cellcolor{lavender}TNT (ours) & \cellcolor{lavender}23.71 & \cellcolor{lavender}3.18 & \cellcolor{lavender}1700 & \cellcolor{lavender}Rate=[.5, .7, .6] \\
        \hline

        \multirow{3}{*}{\rotatebox[origin=c]{90}{$\tiny\text{GFLOPs } \text{\tiny $ \approx 3.10$}$}}
        & Zero-TP~\cite{wang2024zero} & 0.64 & 3.17 & 1179 & Rate=[1., .3, .3, .4, 1.] \\
        & ToMe~\cite{bolya2022token} & 6.30 & 3.26 & 1485 & $r$=25 \\
        & DynamicViT~\cite{rao2021dynamicvit} & 10.36 & 3.53 & 1723 & $\rho$=.2 \\
        & \cellcolor{lavender}TNT (ours) & \cellcolor{lavender}13.42 & \cellcolor{lavender}2.98 & \cellcolor{lavender}1786 & \cellcolor{lavender}Rate=[.5, .6, .55] \\
        \hline
    \end{tabular}

    \caption{
    \textbf{Multi-layer pruning} for ViT/16. in low-token regime.
    }
    \label{tab:vit16_multi_layer_low}
\end{table*}

\section{More examples for visualization}

\begin{figure*}[h!]
    \centering
    \includegraphics[width=\textwidth]{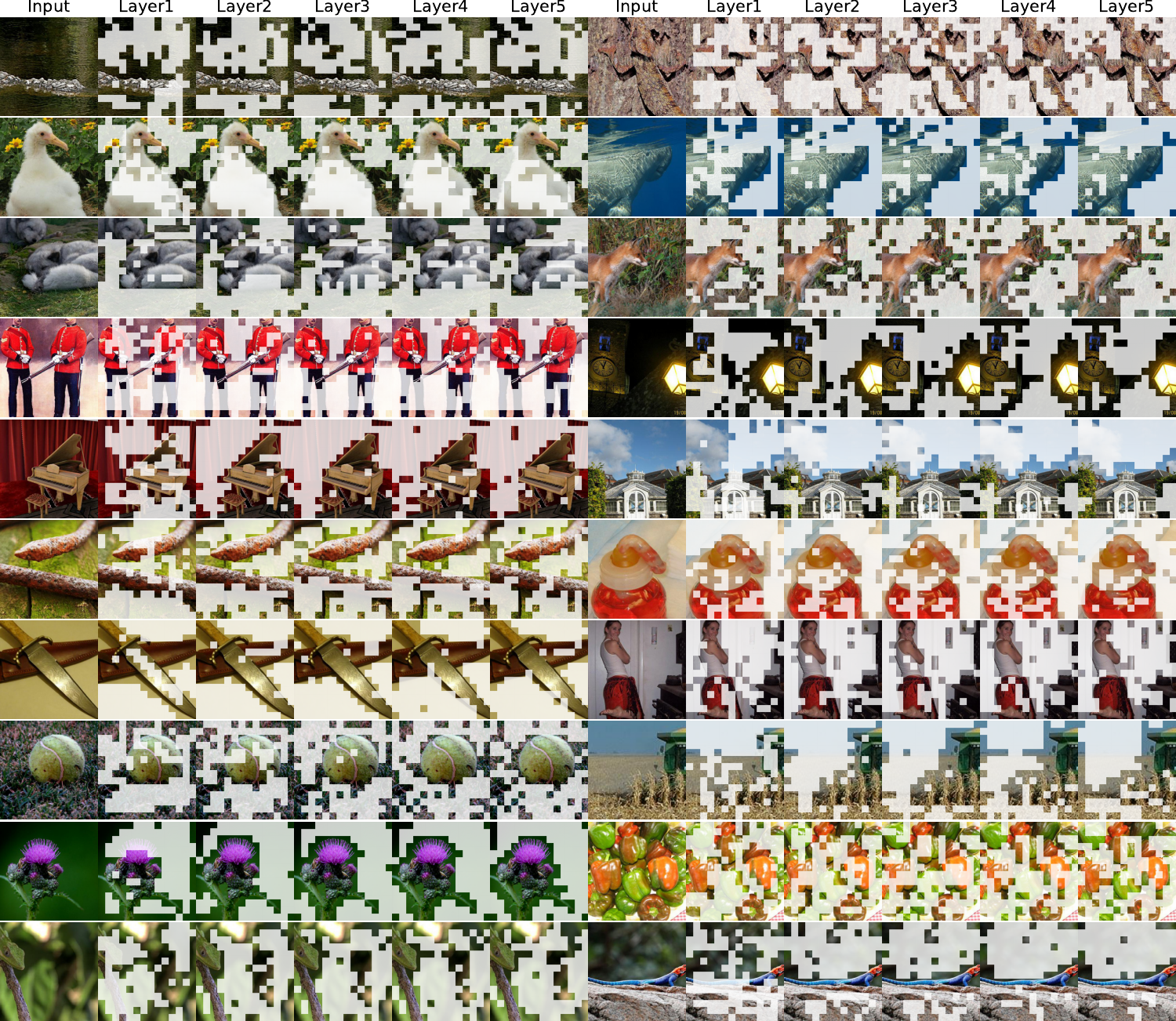}  
    \caption{More Visualization of Token Pruning maps on ImageNet-1K: at \textbf{left} are the original images, and at each column \textbf{progressing right} are single layer prunings and their associated kept/dropped tokens, for layers 1-5 of the DeiT-S-Distil. model.}
    \label{fig:more_visualization}
\end{figure*}

More qualitative results for pruning $50\%$ of tokens at varying layers (single-layer pruning at layers 1-5) on the ImageNet-1K validation dataset in Figure ~\ref{fig:more_visualization}.

\end{document}